\documentclass[journal]{IEEEtran}

\usepackage{import}
\usepackage{tikz}
 \usepackage{cite}

\ifCLASSINFOpdf
  \graphicspath{{./}}
  \DeclareGraphicsExtensions{.pdf,.jpeg,.png}
\else
  \usepackage[dvips]{graphicx}
  \DeclareGraphicsExtensions{.eps}
\fi

\ifCLASSOPTIONcompsoc
  \usepackage[caption=false,font=normalsize,labelfont=sf,textfont=sf]{subfig}
\else
  \usepackage[caption=false,font=footnotesize]{subfig}
\fi

\ifCLASSOPTIONcaptionsoff
  \usepackage[nomarkers]{endfloat}
 \let\MYoriglatexcaption\caption
 \renewcommand{\caption}[2][\relax]{\MYoriglatexcaption[#2]{#2}}
\fi

\usepackage{fixltx2e}
\usepackage{url}
\usepackage{algorithm}
\usepackage{algpseudocode}
\usepackage{mathtools}
\usepackage{amsmath}
\usepackage{makecell}
\usepackage{booktabs}
\usepackage{lipsum}
\usepackage{pgfplots}
\usepackage{textcomp}
\usepackage{siunitx}
\usetikzlibrary{calc}
\usetikzlibrary{patterns}
\usetikzlibrary{chains}
\usetikzlibrary{matrix}
\usepackage{graphicx}
\usepackage{color,soul}
\usepackage{balance}
\usepackage{hyperref}

\algnewcommand\algorithmicforeach{\textbf{for each}}
\algdef{S}[FOR]{ForEach}[1]{\algorithmicforeach\ #1\ \algorithmicdo}

\begin{document}

\title{Combined Static and Motion Features for Deep-Networks Based Activity Recognition in Videos }

\author{Sameera~Ramasinghe,~\IEEEmembership{Member,~IEEE,}
        Jathushan~Rajasegaran,~\IEEEmembership{Student Member,~IEEE,}
        Vinoj~Jayasundara,~\IEEEmembership{Student Member,~IEEE,}
        Kanchana~Ranasinghe,~\IEEEmembership{Student Member,~IEEE,}
        Ranga~Rodrigo,~\IEEEmembership{Member,~IEEE,}
        and~Ajith~A.~Pasqual,~\IEEEmembership{Member,~IEEE}

\thanks{S. Ramasinghe, J. Rajasegaran, V. Jayasundara, K. Ranasinghe, R. Rodrigo, and A. Pasqual were with the Department
of Electronic and Telecommunication Engineering, the University of Moratuwa, 10400, Sri Lanka.
E-mail: samramasinghe@gmail.com, brjathu@gmail.com, vinojjayasundara@gmail.com, kahnchana@gmail.com, ranga@uom.lk, pasqual@ent.mrt.ac.lk}

\thanks{This paper has supplementary downloadable material available at http://ieeexplore.ieee.org, provided by the author. Digital Object Identiﬁer XX.XXXX/TCSVT.2017.XXXXXXX}}

\ifCLASSOPTIONpeerreview
\else
\markboth{ IEEE Transactions On Circuits And Systems for Video Technology, Vol. XX, No. XX, Month 201X}%
{Ramasinghe \MakeLowercase{\textit{et al.}}: Static and Motion Features for Deep-Networks Based Activity Recognition in Videos}
\fi

\IEEEpubid{\begin{minipage}{\textwidth}\ \\[12pt] \centering
  0000--0000/00\$00.00~\copyright~2017 IEEE. Personal use is permitted, but republication/redistribution requires IEEE permission.\\
  See http://www.ieee.org/publications standards/publications/rights/index.html for more information.
\end{minipage}}

\maketitle

\begin{abstract}
    Activity recognition in videos in a deep-learning setting---or otherwise---uses both static and pre-computed motion components. The method of combining the two components, whilst keeping the burden on the deep network less, still remains uninvestigated. Moreover, it is not clear what the level of contribution of individual components is, and how to control the contribution. In this work, we use a combination of CNN-generated static features and motion features in the form of motion tubes. We propose three schemas for combining static and motion components: based on a variance ratio, principal components, and Cholesky decomposition. The Cholesky decomposition based method allows the control of contributions. The ratio given by variance analysis of static and motion features match well with the experimental optimal ratio used in the Cholesky decomposition based method. The resulting activity recognition system is better or on par with existing state-of-the-art when tested with three popular datasets. The findings also enable us to characterize a dataset with respect to its richness in motion information.
\end{abstract}


\section{Introduction}
\tikzset{ font={\fontsize{4pt}{5}\selectfont}}

\IEEEPARstart{A}{utomatic} activity recognition in videos is
an intensely researched area in computer vision due to its wide range of real-world applications in
sports, health care, surveillance, robot vision, and human-computer interaction.
Furthermore, the rapid growth of digital video data demands automatic
classification and indexing of videos. Despite the increased interest, the state-of-the-art
systems are still far from human-level performance, in contrast to the success in image classification \cite{girshick2014rich, krizhevsky2012imagenet}. This is partially due to the complex intra-class variations in videos, some obvious causes being the view point, background
clutter, high dimensionality of data, lack of large datasets, and low resolution.

Despite these reasons more-or-less affecting automatic image classification, it has been quite successful in recent years,
largely owing to the rise of deep learning techniques. This is not the case in video classification, although deep learning is
starting to be widely applied. Therefore, it is worthwhile investigating
what is holding back video classification. In this study, we address three key
areas: exploiting the underlying dynamics of sub-events for high-level action recognition, crafting
self-explanatory, independent static and motion features---in which, motion features should capture micro-level actions
of each actor or object independently, such as arm or leg movements---, and optimum fusing of static and motion features for better accuracy.

A complex activity typically comprises several sub activities.
The existing approaches try to classify a video treating it as a
single, high-level activity \cite{wang2011action, wang2013action, simonyan2014two, 7486474}.
As the action becomes complex, the behavior and the
temporal evolution of its underlying sub-events become complicated. For example,
cooking may involve sub-events: cutting, turning on the cooker, and stirring. It may not always preserve this
same lower order for the same action class. Instead, they may contain
a higher-order temporal relationship among them. For example, turning on the cooker and cutting
may appear in reverse order in another video in the same class.
Therefore, this temporal pattern of sub-events is not easy to capture through a simple
time series analysis. These patterns can only be identified by observing many examples,
using a system which has an infinite dynamic response. Therefore, it is important to
model this higher-order temporal
progression, and capture temporal dynamics of these sub-events for better recognition of
complex activities.

\IEEEpubidadjcol
In contrast to image classification, the information contained in videos
are not in a single domain. Both the motion patterns of the actors and objects, as well as the static
 information---such as, background and still objects that the actors interact with---are important for determining an action.
For example, the body movements of a group of people fighting may closely relate to
the body movements of a sports event, e.g., wrestling. In such a case, it is tough to distinguish between the two
activities solely by looking at the motion patterns. Inspecting the background setting
and objects is crucial in such a scenario. Therefore, it is necessary to engineer powerful features
from both motion and static domains. In addition, these features must be complementary, and one feature domain should
not influence the other domain. In other words, they should be self-explanatory, and mutually exclusive, as much as possible.
Also, these motion features should be able to capture
activities of each actor independently. If so, the distributional properties of these actions, over short
and long durations, can be used to identify high level actions. In the proposed method, we craft static and motion features to have this property.

Furthermore, how to optimally combine or fuse these motion and static descriptors remains a challenge for three key reasons: both static and motion information
provide cues regarding an action in a video, the contribution ratio from each domain affects the final recognition accuracy,
and optimum contribution ratio of static and motion information may depend on the richness of motion information in the video.
The combined descriptor should contain the essence of both domains, and should not have a
negative influence on each other. Recently, there has been attempts to answer this question \cite{7486474},\cite{simonyan2014two}. However,
these existing methodologies lack the insight in to how much this ratio affects the final accuracy, and has no control over this ratio.
One major requirement of the fusion method is to have a high-level intuitive interpretation on how much contribution each domain provides to the
final descriptor, and can control level of contribution. This ability makes it possible
to deeply investigate the optimum contribution of each domain for a better accuracy. In this study, we investigate this factor
extensively.

In this work, we focus on video activity recognition using both static and motion information. We address three problem areas: crafting mutually exclusive static and motion features, optimal fusion of the crafted features, and
modeling temporal dynamics of sub-activities for high-level activity recognition. In order to examine the temporal evolution
of sub-activities subsequently, we create video segments with constant overlapping durations. Afterwards, we combine static and motion
descriptors to represent each segment. We propose \textit{motion tubes},
a dense trajectory \cite{wang2011action} based tracking mechanism, to identify and track candidate moving areas
separately. This enables independent modeling of the activities in each moving area.
In order to capture motion patterns, we use histogram oriented optic flows (HOOF) \cite{chaudhry2009histograms}
inside motion tubes. Then we apply a bag-of-words (BoW) method on the generated
features to capture the distributional properties of micro actions, such as body movements, and create high level, discriminative motion features.

Inspired by the power  of object recognition of convolutional neural networks (CNNs), we create a seven-layer deep
CNN, and pre-train it on the popular ImageNet dataset \cite{deng2012imagenet}.
Afterwards, we use this trained CNN to create deep features, to synthesize static descriptors.
Then, using a computationally efficient, yet powerful, mathematical model, we fuse static and motion feature vectors. We propose three such novel methods in this paper: based on
Cholesky decomposition, variance ratio of motion and static vectors, and principal components analysis (PCA). The Cholesky decomposition based model provides the ability to
precisely control the contribution of static and motion domains to the final fused descriptor. Using this intuition, we investigate the
optimum contribution of each domain, experimentally.  The variance ratio based method also provides us this ability, and additionally, lets us
obtain the optimum contribution ratio mathematically. We show that the optimum contribution ratio obtained experimentally using the Cholesky based method,
matches with the ratio obtained mathematically from the variance based method. Furthermore, we show that this optimum contribution may
vary depending on the richness of motion information, and affects the final accuracy significantly.

In order to model the temporal progression of sub events, we feed the fused vectors to
a long short-term memory (LSTM) network. The LSTM network discovers the underlying temporal patterns of the sub events, and classifies high level actions.
We feed the same vectors to a classifier which does not capture temporal dynamics to show that modeling temporal progression of sub events indeed contribute for a better
accuracy. We carry out our experiments on the three popular action data sets, UCF-11 \cite{liu2009recognizing},
Hollywood2 \cite{marszalek2009actions}, and HMDB51 \cite{Kuehne11}.

The key contributions of this paper are as follows:

 \begin{itemize}
  \item We propose an end-to-end system, which  extracts both static and motion information, fuses, and models the
temporal evolution of sub-events, and does action classification.
  \item We propose a novel, moving actor and object tracking mechanism, called \textit{motion tubes},
which enables the system to track each actor or object independently, and model the motion patterns individually over a long time period.
This allows the system to model actions occurring in a video in micro level, and use these learned dynamics
at a high level afterwards.
 \item We propose three novel, simple, and efficient mathematical models for fusing two vectors,
in two different domains, based on Cholesky transformation, variance ratio of motion and static vectors, and PCA. The first two methods provide
the ability to govern the contribution of each domain for the fused vector precisely and find the optimum contribution of the two domains, mathematically or empirically. Using this advantage, we prove that static and motion information are complementary and vital for activity recognition through experiments.
 \item We prove that the final recognition accuracy depends on the ratio of contribution of static and motion domains. Also, we show that
 this optimum ratio depends on the richness of motion information in the video. Hence, it is beneficial to exploit this optimum ratio for a better accuracy.
  \item We model the underlying temporal evolution of sub-events for complex activity recognition using an LSTM network. We experimentally
prove that capturing these dynamics indeed benefits the final accuracy.
 \end{itemize}

With the proposed techniques we outperform the existing best results for the datasets UCF-11 \cite{liu2009recognizing}
and Hollywood2 \cite{marszalek2009actions}, and are on par for the dataset HMDB51 \cite{Kuehne11}.

\section{Related work}

There has not been many approaches in activity recognition, which highlight the
importance of exclusively engineered static and motion features. Most of the work
rely on generating spatio-temporal interest regions, such as, action tubes \cite{gkioxari2015finding},
tubelets \cite{jain2014action}, dense trajectory based methods \cite{van2015apt, wang2015action},
spatio-temporal extensions of spatial pyramid pooling \cite{laptev2008learning},
or spatio-temporal low-level features  \cite{schuldt2004recognizing, ke2005efficient,shechtman2005space, wang2011action, klaser2008spatio, yu2010real}. Action tubes \cite{gkioxari2015finding} is quite similar to
our motion tubes, but our motion candidate regions are chosen based on more powerful dense trajectories \cite{wang2011action} instead of
raw optic flows. Also, we employ a tracking mechanism of each moving area through motion tubes isolating actions
of each actor throughout the video. This is an extension of the human-actor tracking presented by Wang \cite{wang2013action} . Our static interest regions are independent from motion,
unlike in Gkioxari \textit{et al.}~\cite{gkioxari2015finding}, where can extract background scenery information using CNNs, for
action recognition.
A common attribute of these methods is that motion density is the
dominant factor for identifying candidate regions.
In contrast, we treat motion and static features
as two independent domains, and eliminate the dominance factor.

A few attempts has recently been made on exclusive crafting and late fusion
of motion and static features. Simonyan \textit{et al.}~\cite{simonyan2014two} first decomposes a video in to
spatial and temporal components based on RGB and optical flow frames.
Then they apply two deep CNNs on these two components separately to extract spatial and
temporal information. Each network operates mutually exclusively and performs action classification
independently. Afterwards, softmax scores are coalesced by late fusion.

Work done in Feichtenhofer \textit{et al.}\cite{feichtenhofer2016convolutional} is also similar. Instead of late fusion,
they fuse the two domains in a convolutional layer. Both these approaches rely explicitly on automatic feature
generation in increasingly abstract layers. While this has provided promising results on static feature generation,
we argue that motion patterns can be better extracted by hand-crafted features. This is because
temporal dynamics extend to a longer motion duration unlike spatial variations. It is not possible
to capture and discriminate motion patterns in to classes by a system which has a smaller temporal support. There are models
which employ 3D convolution \cite{ji20133d, tran2015learning}, which extends the traditional CNNs into temporal domain.
Ramasinghe \textit{et al.}\cite{7486474} apply CNNs on optic flows, and Kim \textit{et al.}~\cite{kim2007human} on low level hand-crafted inputs
(spatio-temporal outer boundaries volumes), to extract motion features. However, even by generating hierarchical
features on top of pixel level features, it is not easy to discriminate motion classes as the duration extent is short.
Also, tracking and modeling actions of each actor separately in longer time durations is not possible with these
approaches. Our motion features, on the other hand, are capable of capturing motion patterns in longer temporal durations.
Furthermore, with the aid of \textit{motion tubes} our system tracks and models the activities of each moving area separately.

In the case of work done by Wang \textit{et al.}~\cite{Wang2014ActionRA}, their use of IDT features as motion descriptors and CNN features as static descriptors serves as a baseline for our work. Considering their experiments and observations, we focus solely on HOOF creation as opposed to multiple different descriptors for extracting the motion information. In addition, we look into the extraction of micro actions through our work. Also with regards to the approach for static and motion vector fusion, which involves a constant weighing factor, our work explores alternative approaches with focus on three different methods used across all experimenting. In addition, we improve with regards to capturing the temporal evolution of videos using recurrent neural networks, considering the shortcomings of SVM classifiers in capturing temporal evolutions.

Regarding video evolution, Fernando \textit{et al.}~\cite{fernando2015modeling} postulate
a function capable of ordering the frames of a video
temporally. They learn a ranking function per video using a ranking machine and use the learned parameters as
a video descriptor. Several other methodologies, e.g., HMM \cite{wang2011hidden, wu2014leveraging},
CRF-based methods \cite{song2013action}, also have been employed in this aspect. These methods model the video evolution in frame
level. In contrast, attempts for temporal ordering of atomic events also has been made \cite{rohrbach2012script, bhattacharya2014recognition}.
Rohrbach \textit{et al.}\cite{rohrbach2012script}, encode transition probabilities of a series of events statistically with a HMM model.
Bhattacharya \textit{et al.}\cite{bhattacharya2014recognition} identify low level actions using dense trajectories and assign concept identity
probabilities for each action. They apply a LDS on these generated concept vectors to exploit temporal
dynamics of the low level actions. Li \textit{et al.}\cite{li2013recognizing} uses simple dynamical systems \cite{jackson1992perspectives},
\cite{kailath1974view} to create a dictionary of low-level spatio-temporal attributes. They use these attributes
later as a histogram to represent high level actions. Our method too follows a similar approach,
as we also generate descriptors for sub-events and then extract temporal progression
of these sub-events. However, instead of a simple statistical model, which has a finite dynamic response,
we use an LSTM network \cite{hochreiter1997long} to capture these dynamics. In action recognition literature,
such models are starting to appear. In Yue-Hei \textit{et al.}
\cite{yue2015beyond} the LSTM network models the dynamics of the CNN activations, and in Donahue \textit{et al.}\cite{donahue2015long},
the LSTM network learns the temporal dynamics of low level features generated by a CNN.

\section{Methodology}

\subsection{Overview}

\begin{figure*}
  \centering
\begin{tikzpicture}[x=0.7cm, y=0.7cm, every node/.append style={text=black, font=\scriptsize}]

	  \matrix (m) [matrix of nodes,
    column sep=0.2cm,
    row sep=-0.2cm,
    nodes={draw, 
      anchor=center,
      text centered,
      minimum height=8mm
    },
    right iso/.style={isosceles triangle,scale=0.5,sharp corners, anchor=center, xshift=-4mm},
    left iso/.style={right iso, rotate=180, xshift=-8mm},
    txt/.style={text width=1.5cm,anchor=center},
    ellip/.style={ellipse,scale=0.5},
    empty/.style={draw=none}
    ]
  {
   &   \begin{tikzpicture} \draw (-0.1,-0.3) rectangle ++(0.5,0.5); \draw (0.6,0.6) rectangle ++(0.4, -0.4);
  \draw (0.8,0.3) rectangle ++(0.4, -0.4);
  \draw (1.2,-0.3) rectangle ++(0.4, -0.4);
  \draw (2, 0.3) rectangle ++(0.3, -0.8);
  \draw[-latex] (0.4, -0.05) -- ++(0.3, 0);
  \end{tikzpicture} & Addition &  &
  \begin{tikzpicture}
  	\draw[-latex] (0,0) -- (2.5,0);
  	\draw[-latex] (0,0) -- (0,2);  	
  	\foreach \i/\h/\c in {1/0.6/{rgb:red,4;green,2;yellow,1},2/0.8/{rgb:red,1;green,2;yellow,1},3/1.2/{rgb:red,4;green,1;yellow,1},4/1.5/{rgb:red,1;green,1;blue,4},5/0.8/{rgb:red,4;magenta,4;yellow,1},6/0.5/{rgb:red,1;green,1;yellow,3}}
  	{
  		\draw[fill={\c}] (0.3*\i, 0) rectangle ++(0.3, \h);
  	}
  \end{tikzpicture}
  &  &  & \\
$v_{i,1}$ & & & & & Fusion layer&

  \begin{tikzpicture}
  	\draw[-latex] (0,0) -- (2.5,0);
  	\draw[-latex] (0,0) -- (0,2);  	
  	\foreach \i/\h/\c in {1/0.56/{rgb:red,4;green,2;yellow,1},2/0.72/{rgb:red,1;green,2;yellow,1},3/1.32/{rgb:red,4;green,1;yellow,1},4/1.42/{rgb:red,1;green,1;blue,4},5/0.76/{rgb:red,4;magenta,4;yellow,1},6/0.42/{rgb:red,1;green,1;yellow,3}}
  	{
  		\draw[fill={\c}] (0.3*\i, 0) rectangle ++(0.3, \h);
  	}
  \end{tikzpicture}

 \\
&
\begin{tikzpicture}
	\draw[fill=red!40] (0,0) coordinate (a) -- ++(0.2, -0.1) coordinate (b) -- ++(0.1, 0.15) coordinate (c) -- ++(-0.2, 0.1) coordinate (d) -- cycle;
	\draw[fill=blue!40]  (0,-0.5) coordinate (i) -- ++(0.2, -0.1) coordinate (j) -- ++(0.1, 0.15) coordinate (k) -- ++(-0.2, 0.1) coordinate (l) -- cycle;
	\begin{scope}[xshift=50, yshift=5]
		\draw[fill=red!60]  (0,0) coordinate (e) -- ++(0.2, -0.1) coordinate (f) -- ++(0.1, 0.15) coordinate (g) -- ++(-0.2, 0.1) coordinate (h) -- cycle;
		\draw[fill=blue!60]  (0,-1) coordinate (m) -- ++(0.2, -0.1) coordinate (n) -- ++(0.1, 0.15) coordinate (o) -- ++(-0.2, 0.1) coordinate (p) -- cycle;		
	\end{scope}
	\draw[dashed] (a) .. controls (0.5,0.5) ..  (e);
	\draw (b) .. controls (0.7,0.4) ..  (f);
	\draw (c) .. controls (0.8,0.55) ..  (g);
	\draw (d) .. controls (0.6,0.65) ..  (h);
	
	\draw[dashed] (i) .. controls (0.5,-0.3) ..  (m);
	\draw (j) .. controls (0.7,-0.4) ..  (n);
	\draw (k) .. controls (0.8, -0.25) ..  (o);
	\draw (l) .. controls (0.6, -0.15) ..  (p);	
\end{tikzpicture}
& \begin{tikzpicture}
	\draw (0,0) circle (1);
	\foreach \a in {0, 1, 2, 3, 4, 5}
	{
		\draw (\a*30:1) -- ++(\a*30:-2);
	}
\end{tikzpicture} &
BoW &
  \begin{tikzpicture}
  	\draw[-latex] (0,0) -- (2.5,0);
  	\draw[-latex] (0,0) -- (0,2);  	
  	\foreach \i/\h/\c in {1/0.5/{rgb:red,4;green,2;yellow,1},2/0.6/{rgb:red,1;green,2;yellow,1},3/1.5/{rgb:red,4;green,1;yellow,1},4/1.3/{rgb:red,1;green,1;blue,4},5/0.7/{rgb:red,4;magenta,4;yellow,1},6/0.3/{rgb:red,1;green,1;yellow,3}}
  	{
  		\draw[fill={\c}] (0.3*\i, 0) rectangle ++(0.3, \h);
  	}
  \end{tikzpicture}
\\

$v_{i,n}$ & & & & & & & &

\begin{tikzpicture}[scale=0.3]
	\draw  (0,0) -- ++(2, 0) -- ++(0, 0.5) coordinate  (a)  -- ++(-4, 0) coordinate  (b) -- ++(0, -0.5) -- cycle;

	\draw  (0,2) -- ++(3, 0) coordinate  (c) -- ++(0, 0.5) coordinate (d)  -- ++(-6, 0) coordinate  (e) -- ++(0, -0.5)  coordinate  (f)  -- cycle;
	
	\draw  (0,4) -- ++(1.5, 0)  coordinate  (g) -- ++(0, 0.5)  -- ++(-3, 0) coordinate[midway] (t) -- ++(0, -0.5) coordinate  (h) -- cycle;	
	
	\draw[-latex] (a) -- (f);
	\draw[-latex] (b) -- (c);
	
	\draw[-latex] (e) -- (g);
	\draw[-latex] (d) -- (h);	
	
	\draw[-latex] (t) -- ++(0, 1);		
	\draw[-latex] (e)  .. controls ++(-2,2) and ++(-2, -2) ..  (f);
\end{tikzpicture}
\\
  };

  \node at (m-1-2) [anchor=south, yshift=20] {CNN};
  \node at (m-3-3) [anchor=south, yshift=20] {HOOF};
    \node at (m-4-9) [anchor=south, yshift=20] {LSTM network};
        \node at (m-1-5) [anchor=south, yshift=20] {Static vector};
        \node at (m-3-5) [anchor=south, yshift=20] {Motion vector};
        \node at (m-3-2) [anchor=south, yshift=20] {Motion tubes};

  \draw[-latex] (m-2-1) |- (m-1-2);
  \draw[-latex] (m-2-1) |- (m-3-2);

    \draw[-latex] (m-1-2) -- (m-1-3);
  \draw[-latex] (m-3-2) -- (m-3-3);

   \draw[-latex] (m-1-3) --(m-1-5);
  \draw[-latex] (m-3-3) -- (m-3-4);
    \draw[-latex] (m-3-4) --  (m-3-5);

     \draw[-latex] (m-1-5) -| (m-2-6);
    \draw[-latex] (m-3-5) -|  (m-2-6);

\draw[-latex] (m-2-6) -- (m-2-7);

\draw[-latex] (m-2-7) |- ($(m-4-9.west)+(0, 0.8)$);
\draw[-latex] ($(m-4-9.west)+(-2.4, 0.4)$) -- ++(2.4, 0);
\draw[-latex] ($(m-4-9.west)+(-2.4, 0)$) -- ++(2.4, 0);
\node at ($(m-4-9.west)+(-2.4, -0.25)$) {$\vdots$};
\draw[-latex] ($(m-4-9.west)+(-2.4, -0.8)$) -- ++(2.4, 0);

\node at ($(m-4-1)+(0, 1.5)$) {$\vdots$};
\node at ($(m-4-1)+(5, 0)$) {$\cdots$};

\draw[-latex] (m-4-9) -- ++(2,0) node [anchor=west] {Classification};

\end{tikzpicture}
  \caption{\textbf{Overall methodology}. The whole process consists of five major steps: (i) segmenting a video (ii) crafting static features, (iii) crafting motion features,
  (iv) fusing static and motion features, and (v) capturing temporal evolution of sub events. Static and motion features are independent
  and complementary. We generate static features based on  a pre-trained CNN and motion features based on
  motion tubes, and capture the temporal evolution of sub events using an LSTM network.}

 \label{fi:overall}
\end{figure*}
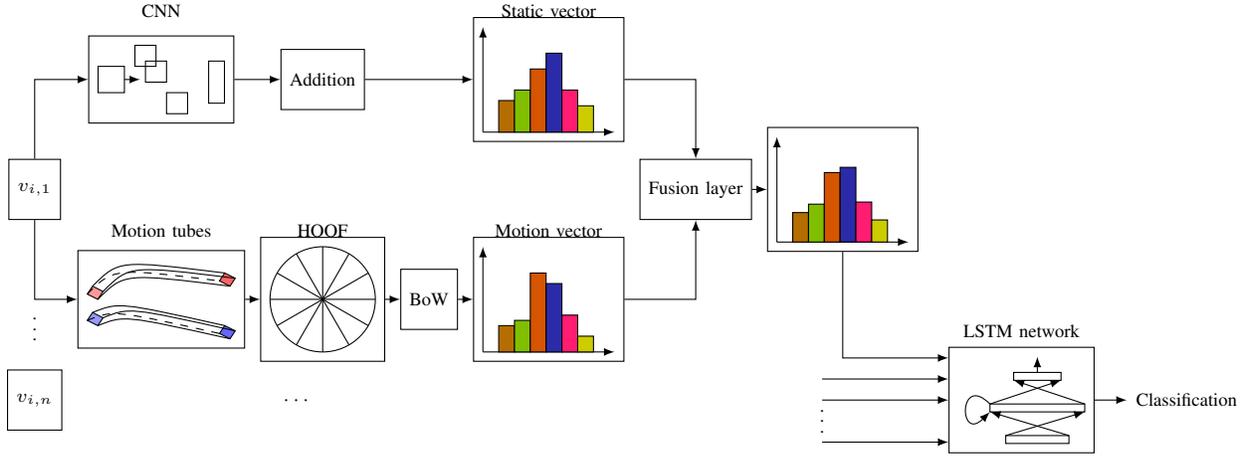

This section outlines our approach. The overall methodology is illustrated in Fig. \ref{fi:overall}.


Our activity classifier classifies video snippets based on their descriptors. In order to compute descriptors, initially, we segment a video into small snippets of 15 frames with a constant frame overlap.
Then we carry out feature construction pipelining for each of these snippets, as shown in Fig. \ref{fi:overall}.
We compute features for each snippet that describe both motion and static domains.
For extracting motion features, we create \textit{motion tubes} across frames, where we track each moving area across the frames using ``action boxes''.
Action boxes are square regions, which exhibit significant motion in each frame. We choose candidate areas by first creating dense trajectories for each frame,
and then clustering trajectory points preceded by a significant amount of pre-processing. This process is explained in sub-section \ref{ss:motion_features}.
These action boxes create motion tubes by getting linking across the frames. Then we calculate HOOF~\cite{chaudhry2009histograms} features within these motion tubes and apply a
bag-of-features method on these to create a motion descriptor for each video segment.

For extracting static features we train a deep CNN on ImageNet. Then we apply this CNN
on the frames of each video snippet to retrieve deep features---output vector from the final softmax layer of the CNN---from it. Then we use these features
to create a static descriptor for the video segment. Afterwards, we combine these motion and static descriptors using one of the fusion models described in sub-section \ref{ss:fusing_static_and_motion}.

The system can then represent a video as a vector time series, $C = [c_{t_0}, c_{t_1}, \dots, c_{t_{n-1}}]$,
where $n$ is the number of segments. Then we apply an LSTM network on these features and exploit the dynamics of time evolution of the combined vector.
Finally, we classify the dynamics of this time series and predict actions.

\subsection{Motion Features}
\label{ss:motion_features}
This section discusses the detailed methodology of creating motion features, particularly, ``Motion tubes'', ``HOOF'', and ``BoW'' blocks as illustrated in Fig. \ref{fi:overall}.

\subsubsection{Low level motion descriptor}
A pixel-level descriptor is required to identify moving points in the video. Dense trajectories \cite{wang2011action} is a powerful
pixel-level algorithm, which captures and tracks motion across several frames. In this work,
we choose dense trajectories as the low-level descriptor for capturing raw motion.

\subsubsection{Clustering}

Initially, we create dense trajectories for every frame in the video.
Then in order to isolate each sub area in a frame which contains significant motion we apply DBScan clustering on the calculated trajectory points.
Algorithm~\ref{alg:Clustering algorithm}  illustrates our clustering approach. Empirically, we use 8 and 10 as $\epsilon$ and MinPoints parameters respectively.

\begin{algorithm}
 \caption{DBScan clustering algorithm.}
   \label{alg:Clustering algorithm}
    \begin{algorithmic}[1]
    \Require  $\mathcal{D}$ \Comment{Dataset of sub-areas in frame}
    \Require $M$ \Comment{Min. no. of points}
    \Require $\epsilon$ \Comment{Max. cluster radius}
        \State $c \leftarrow 0$ \Comment{initialize cluster no.}
        \ForEach {$P \in$ $\mathcal{D}$}
	  \If{$P$ is $visited$}
	    \State{\textbf{continue}}
	  \EndIf
            \State mark $P$ as $visited$
            \State NeighborPts = \textsc{getAllPoints}($P$, $\epsilon$)
          \If{size(NeighborPts) $< M$}
          \State{mark $P$ as noise}
          \Else
          \State{$c$ = next cluster}
          \State{\textsc{addToCluster}($P$, NeighborPts, $c$, $\epsilon$, $M$)}
          \EndIf
        \EndFor

       \Function{addToCluster}{$P$, NeighborPts, c, $\epsilon$, $M$}
	\State add $P$ to cluster $c$
	\ForEach  {point $np \in$ NeighborPts}
	  \If{$P$ is not $visited$}
	    \State mark $np$ as $visited$
	    \State NeighborPts$^\prime$ = \textsc{getAllPoints}($np$, $\epsilon$)
	    \If {size(NeighborPts) $\geq$ $M$}
            \State NeighborPts$^\prime$ $\leftarrow$ NeighborPts joined with NeighborPts$^\prime$
            \EndIf
	
	  \EndIf
	  \If{$np$ is not yet member of any cluster}
         \State add $np$ to cluster $c$
         \EndIf
	\EndFor
       \EndFunction

       \Function{getAllPoints}{$P$, $\epsilon$}
       \Return all points within $P$'s $\epsilon$-neighborhood (including $P$)
       \EndFunction
\end{algorithmic}

\end{algorithm}

Despite the presence of many regions which contain motion in a video, some are neither significant nor descriptive.
Those moving regions can be neglected without loss of performance. Therefore, after clustering each trajectory point in to cluster groups non-significant cluster groups are ignored.
We do this in order to prevent the algorithm focusing on small random moving areas in a video and to limit creation of motion tubes to areas which are significant and descriptive.
Therefore all the clusters which are not at least 50\% the size of the largest cluster of the frame are discarded.

After identifying the initial candidate clusters for creating action boxes, further processing is done to each cluster to ensure that it contains only the important moving areas according to Algorithm~\ref{alg:boundary removal}. For each point, the Chebychev distance from the centroid of the cluster is calculated. 
We discard the furthest 20\% of the points from the cluster. The reason behind the choice of Chebychev distance over Euclidean distance is due to the possibility of obtaining symmetric square shaped cluster groups as opposed to circular ones. This makes it easier to track moving areas and create motion tubes.

\begin{algorithm}
   \caption{Boundary noise removal algorithm of clusters.}
   \label{alg:boundary removal}
    \begin{algorithmic}[1]
    \Require $M_{d}$ \Comment{Max. Chebychev dist.}
    \Require $\mathcal{C}$ \Comment{Input cluster}
	\State {totalPoints $\leftarrow$ points within $M_{d}$ of center of $\mathcal{C}$}
	\State {currentPoints $\leftarrow$ totalPoints}
	\While{$true$}
	  \If{\textsc{count}(currentPoints) $<$ \textsc{count}(totalPoints) $\times 0.8$}
	    \Return {currentPoints}
	  \EndIf
	  \State {$M_{d} \leftarrow M_{d} - 1$}
	  \State {currentPoints $\leftarrow$ points within $M_{d}$ of center of $\mathcal{C}$}	
	\EndWhile

\end{algorithmic}
\end{algorithm}

After identifying square-shaped interest regions (action boxes), we represent each of them with a vector, $b = (x,y,r,f)$, where $x$ and $y$ are the coordinates
of the top left corner of the box, $r$ is the height or width of the box, and $f$ is the frame number.

\subsubsection{Motion Tubes}
Since our work models the time evolution of sub activities within a video, we divide each input video $V_{i}$ into temporal segments, $f(V_{i}) = [v_{i,1},
v_{i,2}, \dots, v_{i,n}]$,
and create features for each individual segment separately. Therefore, after creating the action boxes for each video segment,
the action boxes within a segment $v_{i,t}$ can be represented as,

\begin{equation}
\label{eqn:action box}
\begin{split}
\MoveEqLeft
 g(v_{i,t}) = \big\{[b_{t,1,1},b_{t,1,2},\dots,b_{t,1,q}],\\
 & [b_{t,2,1},b_{t,2,2},\dots,b_{t,2,p}],\dots,[b_{t,n,1},b_{t,n,2},\dots,b_{t,n,k}]\big\}
\end{split}
\end{equation}
where $b_{t,j,k}$ is the $k^{th}$ action box in $j^{th}$ frame of the $t^{th}$ video segment. Note that the number of
action boxes differ from frame to frame.

Therefore, before linking the action boxes, to create motion tubes further pre-processing is needed to ensure the same number of action
boxes exist in every frame within a video segment. Otherwise, different motion tubes could become joined halfway through the video, and the
effort to capture dynamics of each moving object separately is disturbed.

For this purpose, first we calculate the mean number of action boxes per frame in each segment. Then we obtain the rounded down value, $N$, of the mean number. Afterwards, we iterate through each frame starting from frame number $1$ until we come to a frame $W$ which has $N$ number of action boxes. Then from frame $W$ we propagate forward and backward along the frames, either to eliminate or add action boxes. The procedure is explained below. If the action box count in a particular frame is larger than the previous frame, the smallest excess number of action boxes are removed. 

In the case where the action box count is lower than the previous frame, linear regression 
is used for each $x,y$ and $r$ value of vector $b = (x,y,r,f)$
up to that frame, in order to create artificial action boxes until the number of action boxes matches $N$.

Note how this processing results in Eq. \ref{eqn:action box}
being transformed in to Eq. \ref{eqn:action box transformed},
thus verifying that the number of action boxes per frame is equal for all frames within a video segment.

\begin{equation}
\label{eqn:action box transformed}
\begin{split}
\MoveEqLeft
 h(g(v_{i,t})) = \big\{[b_{t,1,1},b_{t,1,2},\dots,b_{t,1,k}],\\
 & [b_{t,2,1},b_{t,2,2},\dots,b_{t,2,k}],\dots,[b_{t,n,1},b_{t,n,2},\dots,b_{t,n,k}]\big\}
\end{split}
\end{equation}

The following procedure is used to link the action boxes in consecutive frames.
Assume $b_{t,k,1}$,$b_{t,k,2}$,\dots,$b_{t,k,n}$ and $b_{t,k+1,1}$,$b_{t,k+1,2}$,\dots,$b_{t,k+1,n}$  are action boxes in two
consecutive frames at time $k$ and $k+1$. Then the following distance matrix is calculated.

\begin{equation}
D=\begin{bmatrix}
    D_{11}       & D_{12} & D_{13} & \dots & D_{1k} \\
    D_{21}       & D_{22} & D_{23} & \dots & D_{2k} \\
    \vdots       & \vdots & \vdots & \vdots & \vdots \\
    D_{k1}       & D_{k2} & D_{k3} & \dots & D_{kk}
\end{bmatrix}
\end{equation}

where $D_{i,j}$ is the Euclidean distance between the centroids of $i^{th}$ action box in $k^{th}$ frame and $j^{th}$ action box in $(k+1)^{st}$ frame.
Then $u^{th}$ action box at $k+1$, and $1^{st}$ action box at $k$ are linked, where $u$ is found using,

\begin{equation}
u=\underset{j\in J}{\mathrm{argmin}_j}\{D_{1,j}\}, J=\{1,2,\dots,l\}
\label{link eq}
\end{equation}

Then the $1^{st}$ row and the $u^{th}$ column are removed from the distance matrix, and we apply the same process repeatedly using Eq. \ref{link eq}
to link each of the action boxes at $k$ with $k+1$.

By this removal process, we avoid combining of motion tubes half-way through
the video segment and keep them isolated from each other, which is vital for capturing the dynamics separately for each moving object.

Finally, we create a $(z\times n)$-by-$5$ matrix $M_{i}$---$z$ and $n$ are number-of-frames and number-of-action-boxes-per-frame, respectively---which
encodes all the information of motion tubes, in a particular video segment $i$. The rows of $M_{i}$ for the $k^{th}$ frame is shown in Eq.~\ref{tubeMatrix},

\begin{equation}
M_{i}=\begin{bmatrix}
    k       & 1 & x_{z,1} & y_{z,1} & r_{z,1} \\
    k       & 2 & x_{z,2} & y_{z,2} & r_{z,2} \\
    \vdots       & \vdots & \vdots & \vdots & \vdots \\
    k       & n & x_{z,n} & y_{z,n} & r_{z,n} \\
\end{bmatrix}
\label{tubeMatrix}
\end{equation}
where the columns represent the frame number, action box number, $x$ coordinate of the top left corner of the action box,
$y$ coordinate of the top left corner of the action box, and the width/height of the action box, respectively.

\subsubsection{Histogram Oriented Optic Flows (HOOF)}
\begin{figure}
  \centering
  \begin{tikzpicture}[x=1cm, y=1cm, every node/.append style={text=black, font=\scriptsize}]
    \def\r{1.5}
    	\draw (0,0) coordinate (o) circle (\r);
    	\foreach \theta in {-90, -60, ..., 240}
    	{
    		\draw (o) -- (\theta:\r);
    	}
    	\foreach \theta/\l in {-60/1, -30/2, 0/3, 30/4, 60/5, 90/6}
    	{
    		\node at  (\theta - 15:1.3) {$\l$};
    	}	
    	\foreach \theta/\l in {-90/1, -120/2, -150/3, -180/4, -210/5, -240/6}
    	{
    		\node at  (\theta - 15:1.3) {$\l$};
    	}		
    	
    	\foreach \theta/\r/\t in {-15/1.5/1, 45/1.2/2, -135/1/3, -165/1.6/4}
    	{
    		\draw[-latex, thick] (o) -- (\theta:\r ) coordinate [near end] (t\t);
    	}
    	
    	\foreach \i/\l in {-1.25/1, -0.75/2, -0.25/3, 0.25/4, 0.75/5, 1.25/6 }
    	{
    		\draw (\i-0.25, -2) rectangle ++(0.5, -0.5);
    		\node (a\l) at (\i, -2.25) {$\l$} ;
    	}
    	
    	\draw[-latex, dashed] (t1) -- (a3);
    	\draw[-latex, dashed]  (t2) -- (a5);
    	\draw[-latex, dashed]  (t3) -- (a2);
    	\draw[-latex, dashed]  (t4) -- (a3);

    \end{tikzpicture}
  \caption{Example HOOF generation with 6 bins}\label{fi:hoof}
\end{figure}
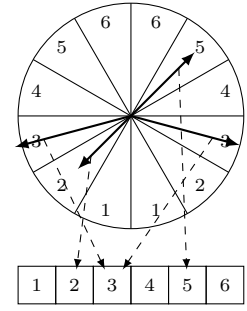

Since each action box in a particular motion tube may differ in size, we take $R = \max(r_{i})$, for $\forall{i}$,
where $r_{i}$ is the length of the $i^{th}$ action box of the motion tube. Then we redraw the action boxes around their centroids having width or length as $R$.
After identifying the $k$ number of motion tubes ($k$ is a variable) for each video segment $v_{i,n}$, we calculate the optic flows along each motion tube
using Lucas \textit{et al.} \cite{lucas1981iterative}.
After that we create HOOF\cite{chaudhry2009histograms} for every
    action box within a motion tube. Each optic flow vector within a spatio-temporal action box within a motion tube is binned according
    to its primary angle from the horizontal axis and weighted according to its magnitude.  Therefore, all optical flow vectors, $z=[x,y]^T$ with direction,
$\theta = \tan^{-1}(\frac{x}{y})$ in the range,

\begin{equation}
- \frac{\pi}{2} + \pi\frac{b-1}{B} \leq \theta < -\frac{\pi}{2} + \pi\frac{b}{B}
\end{equation}

will contribute a weight of $\sqrt{x^2 + y^2}$ to the sum in bin $b$, $1 \leq b \leq B$ out of a total of
$B$ bins. Finally, the histogram is normalized. We choose 100 number of bins. Example HOOF creation for 6 bins is illustrated in Fig. \ref{fi:hoof}.

\subsubsection{Bag of HOOFs}
We use a bag of features method to create a motion descriptor for each video segment. First, we create a code book for HOOF vectors.
100,000 vectors are randomly selected from all the HOOF vectors of all the video segments in all video classes.
Then these 100,000 vectors are clustered using $k$-means clustering and 1000 cluster heads
are identified. We choose the number of cluster heads as 1000, because the dimensions of final motion descriptors are needed to be the same as
the static descriptors, which is explained in section V. Then for each video segment $v_{i,n}$, a histogram is calculated as follows.

We calculate,
\begin{equation}
p = \underset{j\in J}{\mathrm{argmin}_{j}}(T_{j}-h_{n,k}), J=\{1,2,\dots,1000\}
\end{equation}

for each $k$ in $\{1,2,\dots, l\}$, where $h_{n,k}$ is the $k^{th}$ HOOF vector of the $n^{th}$ video segment, and $T_{j}$ is the $j^{th}$ cluster head. Then we increment the histogram values as,

\begin{equation}
H_{n}(p) = H_{n}(p)+1
\end{equation}
where $H_{n}(p)$ is the $p^{th}$ value, $1\leq p\leq 1000$, of histogram of the $n^{th}$ video segment $v_{i,n}$. After calculating the histogram vector $H_{n}$ for every video segment $v_{i,n}$
this $H = [H_{1},H_{2}, \dots, H_{n}]$ is the vector time series, which encodes the time evolution of motion information in the video.

\subsection{Static Features}
In this sub section we discuss crafting of static features. This section relates to
the ``CNN'' and ``Addition'' blocks in Fig. \ref{fi:overall}.

\begin{figure*}
  \centering
  \includegraphics[scale=0.5]{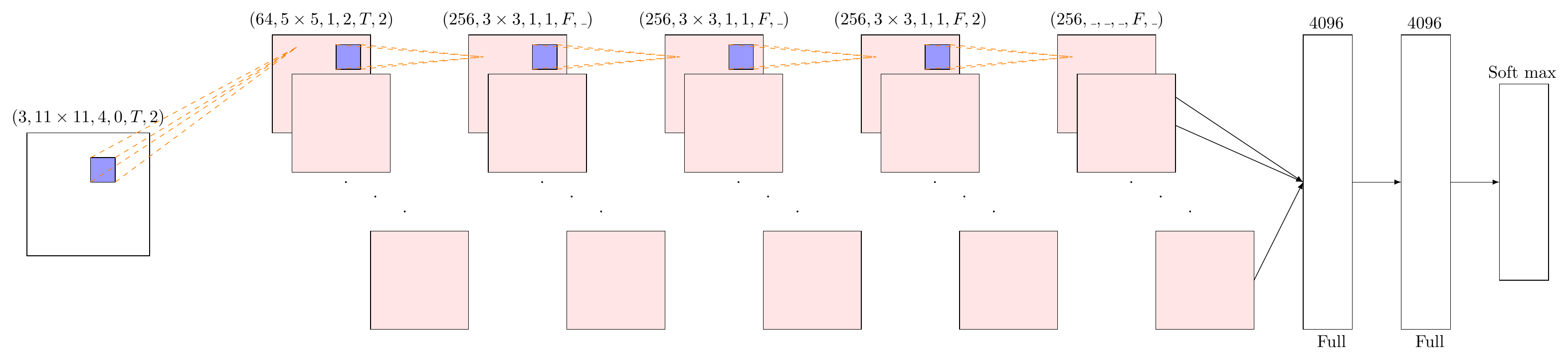}
  \caption{CNN architecture used for generating static features. The CNN consists of five convolution layers,
  two fully connected layers, and one softmax layer. The details of each convolutional layer are provided on top of each layer
  according to the following format:(number of convolution layers $\times$ filter width $\times$ filter height, convolution stride,
  spatial padding, is Local Response Normalization added, max-pooling factor). Value above fully connected layers indicates the dimensionality of the layer.
  We use ReLu as the activation function.}
\label{fi:cnn}
\end{figure*}

In order to create static descriptors, we create a CNN of 1000 output classes, and train it on the ImageNet dataset. The architecture is shown in figure \ref{fi:cnn}.
After training, we apply the trained model on each individual frame of each video segment. Then we average the output vectors of the CNN
along indices and obtain a static descriptor $s_{i}$ for each video segment $v_{i}$. Following the same
procedure for every $v_{i}$, we develop a vector time series,
$S =[s_{1}, s_{2}, \dots, s_{n}]$, representing the static time evolution of the whole video.

\subsection{Fusing of Static and Motion Features}
\label{ss:fusing_static_and_motion}
This work depends on three factors; both static and motion information
are vital for action recognition, but the final accuracy depends on the ratio of contribution of each domain,
and the optimum contribution may depend on the richness of motion information in the video. We derive our fusion models addressing all these aspects.
The three mathematical models we use
to fuse the static and motion vectors are described next. This sub section relates to the ``Fusion layer'' block in Fig. \ref{fi:overall}.

\subsubsection{Cholesky Transformation Based Method}

This derivation is based on the Cholesky transformation. An
abstract version of Cholesky transformation is described below.

Let $P$ and $Q$ be two random variables of unknown correlation. These random variables can be
transformed into two new random variables ($R$ and $S$) with a known correlation of $\rho$, where the
value of $\rho$ can be chosen at will. The transformation can be performed as follows.

\begin{equation}
\begin{bmatrix}
    Y     \\
    Z
\end{bmatrix}
=
\begin{bmatrix}
    1  & 0 \\
    \rho  & \sqrt{1-\rho^2}
\end{bmatrix}
\times
\begin{bmatrix}
    P     \\
    Q
\end{bmatrix}
\end{equation}

Therefore,

\begin{equation}
Y = P
\end{equation}

and

\begin{equation}
Z = \rho P + \sqrt{1-\rho^2}Q
\end{equation}

The Cholesky transformation guarantees that the correlation between the two random variables
$Y$ and $Z$ is $\rho$.

Based on the above properties of the Cholesky transformation, we propose the following methodology to fuse the static and motion vectors.

Let $S$ and $M$ be static and motion vectors, respectively. Cholesky transformation can be applied to the two vectors $S$ and $M$ with the correlation value $\rho_{1}$.

\begin{equation}
\begin{bmatrix}
    Y     \\
    Z
\end{bmatrix}
=
\begin{bmatrix}
    1  & 0 \\
    \rho_{1}  & \sqrt{1-\rho_{1}^2}
\end{bmatrix}
\times
\begin{bmatrix}
    S    \\
    M
\end{bmatrix}
\end{equation}

\begin{equation}
Y = S
\end{equation}

\begin{equation}
Z = \rho_{1} M + \sqrt{1-\rho_{1}^2}M
\end{equation}

Similarly, the transformation can be applied to $M$ and $S$ with the correlation value $\rho_{2}$.

\begin{equation}
\begin{bmatrix}
    A     \\
    B
\end{bmatrix}
=
\begin{bmatrix}
    1  & 0 \\
    \rho_{2}  & \sqrt{1-\rho_{2}^2}
\end{bmatrix}
\times
\begin{bmatrix}
    M    \\
    S
\end{bmatrix}
\end{equation}

\begin{equation}
A = M
\end{equation}

\begin{equation}
B = \rho_{2} S + \sqrt{1-\rho_{2}^2}S
\end{equation}

Again, the Cholesky transformation guarantees the following two properties.

\begin{enumerate}
  \item The correlation between $S$ and $Z$ is $\rho_{1}$.
  \item The correlation between $M$ and $B$ is $\rho_{2}$.
\end{enumerate}

Therefore, if the values of $\rho_{1}$ and $\rho_{2}$ are chosen in such a way that they obey the following
rule,

\begin{equation}
\rho_{2} = \sqrt{1-\rho_{1}^2}
\end{equation}

it can be guaranteed that $Z = B$, $\forall S,M,\rho_{1},\rho_{2}$. Hence, the resultant vector $C$ can be obtained
by,

\begin{equation}
C=Z=B
\end{equation}

where the correlation between $C$ and $S$ is $\rho_{1}$ and the correlation between $C$ and $M$ is $\rho_{1}$. Here $S$ and $M$ represent the static and the
motion vectors whereas $C$ represents the resultant vector. This derivation
leads us to an important intuition: by choosing the value of $\rho_{1}$, we can
choose the degree to which the static features and the motion features contribute,
in deriving the resultant vector. In section 4, it is shown, how this property is used to explore, the optimal contribution of
static and motion domain information for recognizing actions. The derivation of an optimum ratio between these would require either a continuous variation of the ratio for test datasets, or a theoretical derivation of an optimum ratio. However, we hypothesize that this ratio would depend on various characteristics of a given dataset (which we explore over our experiments), and the intention of this study is towards establishing the existence of an optimum ratio, and not the derivation of that optimum ratio.

\subsubsection{Variance Ratio Based Method}

The second method employed to combine the motion and static vectors is based on a Gaussian probability model. We model each vector as a histogram. 
Using the histogram model, the mean and variance of each vector are calculated in order to fit the two vectors into Gaussian distributions. The joint Gaussian distribution is then computed based on this data.

\begin{equation}
G_{sm}(N)= \frac{1}{2\pi\sigma_m\sigma_s} e^{-\left[\frac{[N-\mu_s]^2}{2\sigma_s^2}+ \frac{[N-\mu_m]^2}{2\sigma_m^2} \right]}
\end{equation}

This computation corresponds to the evaluation line obtained when equating the two random variables. This evaluation line is the diagonal through the origin of the static vector vs motion vector plot. The combined distribution is then obtained through this process.

It must be noted that both histograms (corresponding to static and motion components) do not contain equal information. Hence varying the contribution of each histogram to the resultant distribution is necessary. This requires varying of the evaluation line which can be achieved through scaling of the motion and static axes. This scaling process is carried out by the following matrix.

\[\mathrm{Scaling~matrix} =
\begin{bmatrix}
    \frac {\sigma_s}{\sigma_s + \sigma_m} & 0  \\
    0 & \frac {\sigma_m}{\sigma_s + \sigma_m}
\end{bmatrix}
\]

We may conclude that higher variance of a component along one axis reflects lower detail in the model with regards to the other axis. Considering the motion axis, the contribution of this vector towards the resultant vector may be defined by $(1-\frac{\sigma_m}{\sigma_m+\sigma_s})$. A high variance always corresponds to a flatter histogram containing less detail.

This parameter we derive is significant as it defines the contribution of each individual motion and static vector pair independent of explicit terms. Hence the optimum ratio for combination of motion and static components of a given dataset can be mathematically evaluated. With regards to the datasets used for experimenting, 30\% of motion
vector and 70\% of static vector constitute this parameter on average. This mathematical inference is further verified through the experiment results in section IV.

Defining new parameters $N_s$ and $N_s$ as follows, we build a new distribution which is a scaled version of the joint Gaussian distribution obtained previously.

\begin{align*}
N_s &= N \frac{\sigma_m}{\sigma_m+\sigma_s}  \\
N_s &= N \frac{\sigma_s}{\sigma_m+\sigma_s}
\end{align*}

Finally, we define the following distribution representative of the combined vector.

\begin{equation}
G_{sm}(N)= \frac{e^{-\frac{1}{2(1-\rho^2)}\left[\frac{[N_s-\mu_s']^2}{2\sigma_s'^2} + \frac{[N_m-\mu_m']^2}{2\sigma_m'^2} - \frac{2\rho[N_s-\mu_s'][N_m-\mu_m']}{\sigma_s' \sigma_m'}  \right]}}{2\pi\sigma_m'\sigma_s'\sqrt{1-\rho^2}}
\end{equation}

The corresponding mean and variance of this newly computed distribution are denoted by $\mu_s'$, $\sigma_s'$ and $\mu_m'$, $\sigma_m'$ for the static and motion vector respectively.

\subsubsection{PCA Based Method}

The third fusion method is based on Principle Component Analysis (PCA). 
If there are $n$ number of features in the input vector of the PCA, the output of the
PCA will provide a new set of $n$ features which are orthogonal and uncorrelated. Also, if
$\mathrm{output} =[a_{1}, a_{2}, ..., a_{n}]$, then $\mathrm{var}(a_{1})> \mathrm{var}(a_{2})> \dots > \mathrm{var} (a_{n})$.
Due to the properties of the PCA, the original set of n features can be
represented precisely using the first $k (n>k)$ principal
components of the output vector, given that the total squared reconstruction error
is minimized.

In other words, the essence of the original $n$ dimensional dataset is now almost
completely represented by the new $k$ dimensional dataset with a minor data loss. Thus, the
dimension of the dataset is reduced from $n$ to $k$.

In our work, the dimensionality of the dataset $T$ is 2 (number of feature domains: static
and motion).
Using PCA on $T$, we receive a new set $T$', with 2 new features domains. We need only one new feature
domain in the feature space to
represent motion and static domains. Therefore, it is our aim to extract only the first principal component.
In order to do this, we need to justify that the first principal component contains a significant majority of
the essence of the original dataset. In other words, only a negligible amount of data is lost by
eliminating the second principal component.

Therefore, we perform PCA on over 15,000 samples of motion and static
vectors and plot the variance percentage of the total variance
explained by the first and the second principal components respectively, as in figure and figure

\begin{figure}
  \centering
  \includegraphics[width=3.5in]{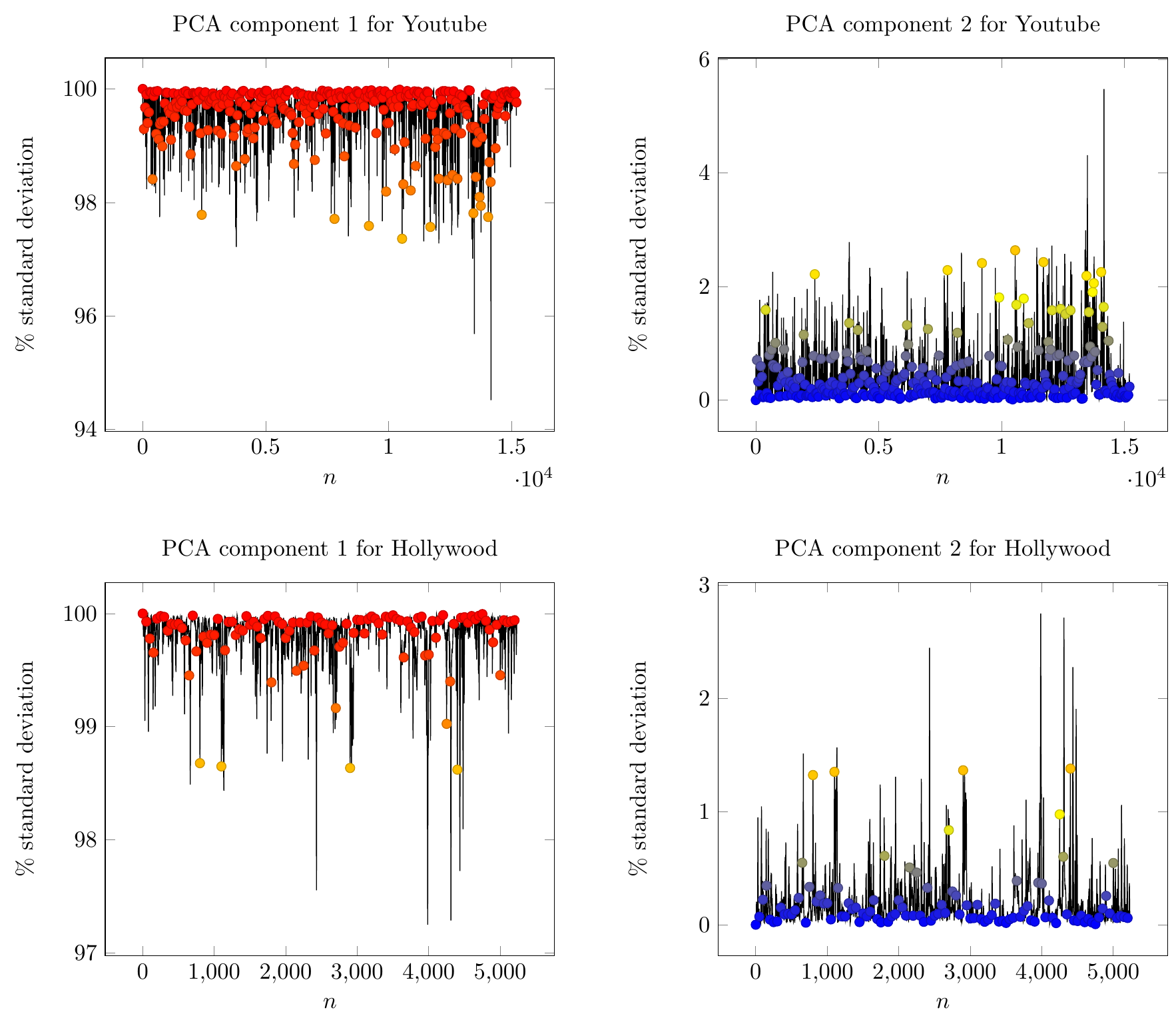}
  \caption{Percent standard deviation values for the first and second components of the PCA. $n$ is the feature vector index. }\label{fi:pca}
\end{figure}

It is evident from the figure that the first principal component almost always accounts
for over 97\% of the essence of the original dataset, except for only a negligible amount
of samples. The lowest percentage registered is approximately 85\%, which is still a
significantly high value. Figure \ref{fi:pca} shows the percent standard deviation values for the first and second components of the PCA for the two datasets.

Therefore, by eliminating the second principal component, only less than 5\% of data
is lost in average, and hence only the first principal component can be used to accurately
represent the original dataset.

\subsection{Capturing Temporal Evolution}
Our work requires analyzing complex dynamic temporal patterns in the generated sequences. This sub section relates to the ``LSTM network'' block shown in Fig. \ref{fi:overall} used for this purpose.

In our work, we represent each video as $n$ fixed-length segments ($n$ differs for videos of different lengths) with overlapping frames.  Each segment is represented with a fused vector $c_{t}$. Therefore, each video can be represented as a vector time series.

Now each vector time series could be analyzed using traditional time series modeling techniques, such as Auto Regressive Moving Average, to obtain features or model parameters that can describe the vector time series. But the main drawback of these methods is that they model current values of a series as a function of past values and have finite dynamic response to time series input. Also, they lack the ability to grasp the internal state representations of a complex time series. RNNs maintain hidden layers with directed feedback connections, and hence, have an infinite dynamic response. While training, it learns internal states of a sequence and usually performs better in modeling complex dynamic temporal patterns of long sequences.

However, it is not ideal to train standard RNNs to solve problems, which require learning of long-term temporal dependencies. This is because of the vanishing-gradient problem, which occurs
due to the exponential decay of gradient loss of the function with respect to time.

In practice, LSTM networks typically perform better in such cases.
LSTM networks are a special type of RNN, which include a ``memory cell'', and as the name suggests,
it can maintain a certain state in memory for longer periods of time.
It also has a set of control gates for controlling the removal or addition of information to the cell state.
This special architecture gives them the ability to capture more long-term dependencies. First, we revise the operation of an LSTM network.

The most important structure of an LSTM unit is its memory cell $c_{t}$, which preserves the state. Basic structure of an LSTM unit is shown in figure \ref{fi:lstmblock}. The memory cell is self-connected, and it has three gates (multiplicative units), i.e., input gate, forget gate and
output gate, which are used to control how much long range contextual information of a temporal sequence to store, remove or output.

The detailed activation process of the memory cell and three gates, as shown in Fig. \ref{fi:lstmblock} is
illustrated as follows:

\begin{figure}
  \centering
  \begin{tikzpicture}[x=0.5cm, y=0.5cm, every node/.append style={text=black, font=\scriptsize}]

	\node [draw, circle, minimum height=18, fill=blue!20] (input) at (0,0) {$i^t$};
	\node [draw, circle, minimum height=18, fill=blue!20] (output) at (6,0) {$o^t$};	
	\node [draw, circle, minimum height=18, fill=red!20] (a) at (-2,-2) {};	
	\node [draw, circle, minimum height=8] (b) at (0,-2) {};
	\node [draw, circle, minimum height=24, fill=green!20] (c) at (2,-2) {$c^t$};	
	\node [draw, circle, minimum height=18, fill=red!20] (d) at (4,-2) {};	
	\node [draw, circle, minimum height=8] (e) at (6,-2) {};	
	\node [] (h) at (8,-2) {$h^t$};	
	\node [draw, circle, minimum height=8] (g) at (2,-3.5) {};	
	\node [draw, circle, minimum height=18, fill=blue!20] (f) at (2,-5) {$f^t$};		
	
	\draw (a.south west) .. controls ( $(a) +(0.2,-0.4)$)  and  ( $(a) +(-0.2,0.4)$) .. (a.north east);
	\draw (d.south west) .. controls ( $(d) +(0.2,-0.4)$)  and  ( $(d) +(-0.2,0.4)$) .. (d.north east);
	\draw (b.south west) -- (b.north east) (b.north west) -- (b.south east);
	\draw (e.south west) -- (e.north east) (e.north west) -- (e.south east);
	\draw (g.south west) -- (g.north east) (g.north west) -- (g.south east);	
	
	\draw[-latex] (a) -- (b);
	\draw[-latex] (b) -- (c);
	\draw[-latex] (c) -- (d);
	\draw[-latex] (d) -- (e);
	\draw[-latex] (e) -- (h);
	\draw[-latex] (output) -- (e);
	\draw[-latex] (c) -- (output);
	\draw[-latex] (c) -- (input);
	\draw[-latex] (input) -- (b);
	\draw[-latex] (f) -- (g);
	
	\draw[-latex] (c)  to[out=-150,in=-180] 	(f);
	
	\draw[-latex] (c.south east)   to[out=-45,in=0] 	node[midway, anchor=west] {Feedback} (g.east) ;
	\draw[-latex] (g.west)   to[out=180,in=-135] 	(c.south west);	
	
	\draw[latex-] (input) -- ++(60:1.5);
	\draw[latex-] (input) -- ++(90:1.5) node[anchor=south] {$X^t$};
	\draw[latex-] (input) -- ++(120:1.5);
	
	\draw[latex-] (output) -- ++(60:1.5);
	\draw[latex-] (output) -- ++(90:1.5) node[anchor=south] {$X^t$};
	\draw[latex-] (output) -- ++(120:1.5);

	\draw[latex-] (a) -- ++(150:1.5);
	\draw[latex-] (a) -- ++(180:1.5) node[anchor=east] {$X^t$};;
	\draw[latex-] (a) -- ++(210:1.5);
	
	\draw[latex-] (f) -- ++(-120:1.5);
	\draw[latex-] (f) -- ++(-90:1.5) node[anchor=north] {$X^t$};;
	\draw[latex-] (f) -- ++(-60:1.5);
	
	\node at (f.east) [anchor=west] {Forget gate};
	\node at (output.west) [anchor=east, text width=0.8cm] {Output gate};
	\node at (input.west) [anchor=east, text width=0.6cm] {Input gate};
	\node at (c.north) [anchor=south] {Cell};

\draw  (-2.8,0.9) rectangle (7,-6);
\end{tikzpicture}
  \caption{Long short-term memory (LSTM) block cell. Source \cite{Graves2008supervised}.}\label{fi:lstmblock}
\end{figure}
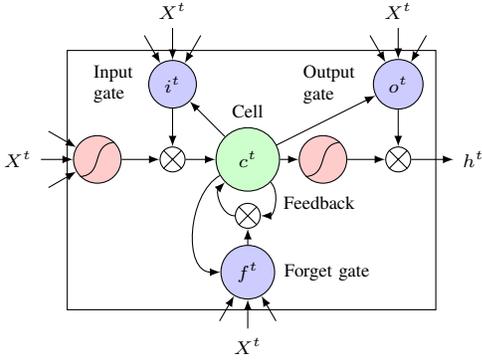

\begin{equation}
i^{t} = \sigma (W_{xi}x^t + W_{hi}h^{t-1} + W_{ci}c^{t-1} + b_{i})
\end{equation}
\begin{equation}
f^{t} = \sigma (W_{xf}x^t + W_{hf}h^{t-1} + W_{cf}c^{t-1} + b_{f})
\end{equation}
\begin{equation}
c^{t} = f^tc^{t-1} + i^ttanh(W_{xc}x^t + W_{hc}h^{t-1} + b_{c})
\end{equation}
\begin{equation}
o^{t} = \sigma (W_{xo}x^t + W_{ho}h^{t-1} + W_{co}c^{t-1} + b_{o})
\end{equation}
\begin{equation}
h^t = o^t\tanh(c^t)
\end{equation}

where $W$ is the connection weight between two units and $\sigma(\cdot)$ is the sigmoid function.

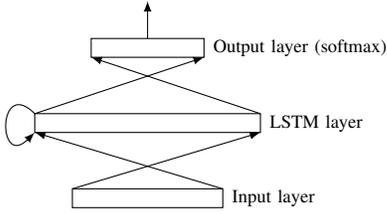
\begin{figure}
  \centering
  \begin{tikzpicture}[x=0.5cm, y=0.5cm, every node/.append style={text=black, font=\scriptsize}]
	\draw  (0,0) -- ++(2, 0) -- ++(0, 0.5) coordinate  (a)  node [midway, anchor=west] {Input layer} -- ++(-4, 0) coordinate  (b) -- ++(0, -0.5) -- cycle;

	\draw  (0,2) -- ++(3, 0) coordinate  (c) -- ++(0, 0.5) coordinate (d) node [midway, anchor=west] {LSTM layer}  -- ++(-6, 0) coordinate  (e) -- ++(0, -0.5)  coordinate  (f)  -- cycle;
	
	\draw  (0,4) -- ++(1.5, 0)  coordinate  (g) -- ++(0, 0.5) node [midway, anchor=west] {Output layer (softmax)}  -- ++(-3, 0) coordinate[midway] (t) -- ++(0, -0.5) coordinate  (h) -- cycle;	
	
	\draw[-latex] (a) -- (f);
	\draw[-latex] (b) -- (c);
	
	\draw[-latex] (e) -- (g);
	\draw[-latex] (d) -- (h);	
	
	\draw[-latex] (t) -- ++(0, 1);		
	\draw[-latex] (e)  .. controls ++(-1,1) and ++(-1, -1) ..  (f);
\end{tikzpicture}
  \caption{A simple illustration of the LSTM network. The network consists of an input layer, a 128-unit LSTM layer with 0.8 dropout, and
a fully-connected softmax output layer.}\label{fi:layers}
\end{figure}

\begin{figure}
  \centering
  \begin{tikzpicture}[x=0.7cm, y=0.7cm, every node/.append style={text=black, font=\scriptsize}]
	\foreach \i/\l in {1/1, 2/2, 3/3, 4/4, 6/n}
	{
		\node[draw=black, minimum height=20] (1\i) at (2*\i, 0) {LSTM};
		\node[draw=black, minimum height=20, align=center] (2\i) at (2*\i, -1.5) {$w_l$};
		\node[text width=30, align=center] (3\i) at (2*\i, -3) {$c_\l$};		
	}
	
	\node (15) at (2*5,0) {$\cdots$};
	\node (25) at (2*5,-1.5) {$\cdots$};
	\node (35) at (2*5,-3) {$\cdots$};	
	
	\foreach  \i/\j in {1/2, 2/3, 3/4, 4/5, 5/6}
	{
		\draw [-latex] (1\i) -- (1\j);
	}
	
	\foreach  \i in {1,2,3,4,6}
	{
		\draw [-latex] (2\i) -- (1\i);
		\draw [-latex] (3\i) -- (2\i);
	}	
	
	\draw[-latex] (16) -- +(0,1) node [anchor=south] {Output};

\end{tikzpicture}
  \caption{The process of feeding fused vectors to the LSTM network. $c_{i}$ indicates the fused vector representing the $i_{th}$
  video segment.}\label{fi:lstm}
\end{figure}
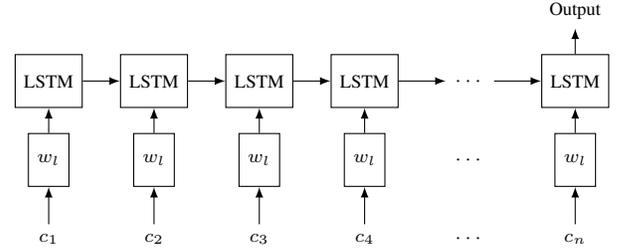

Since the LSTM network is used only for capturing the temporal dynamic patterns between sub actions, one LSTM layer is enough.
Our LSTM network is shown in Fig. \ref{fi:layers}. The network consists of an input layer, a 128-unit LSTM layer with 0.8 dropout, and
a fully connected softmax output layer. As we have a sequence of activities per classification, we use a many-to-one approach
for feeding the fused vectors to the network, as shown in Fig. \ref{fi:lstm}.

\section{Experiments and Results}
This section details our experimental methodology and the  video  datasets used. We evaluate our approach on the two popular datasets
UCF-11 and Hollywood2. On both datasets, we show that our work exceeds the current state-of-the-art
results. We also vary the contribution of static and motion features for the calculation
of combined vector series and explore what is optimum contribution from each domain. We show that optimum contribution
may vary depending on the dataset. We also show that static and motion features
are complementary, and provide vital information about the actions occurring in a video. We compare our three fusion models and show that
all the methods are better or on par with existing state-of-the-art. Furthermore, we highlight the importance of considering the time evolution
of sub activities in order to identify complex events by comparing the results of LSTM and Random Forest Classification algorithm(which does not capture
the temporal dynamics), when applied on our features.

\subsection{Datasets}
\noindent
\textbf{Holywood 2} \cite{marszalek2009actions}: This consists of 12 classes of human actions distributed over 1500 video clips:
\textit{answer phone, drive car, eat, fight person, get out car, hand shake,
hug person, kiss, run, sit down, sit up, }and \textit{stand up}.
The dataset is composed of video clips from 69 movies and provides a challenging task, in automatic action detection.

\noindent
\textbf{UCF-11} \cite{liu2009recognizing}: This consists over 1000
sports and home videos from YouTube. This dataset contains 11 action classes:
\textit{basketball shooting, cycle, dive, golf swing, horse
back ride, soccer juggle, swing, tennis swing, trampoline
jump, volleyball spike}, and \textit{walk with a dog}. Each of the action
sets is subdivided into 25 groups sharing similar environment conditions.
This is a challenging dataset with
camera jitter, cluttered backgrounds and variable illumination.

\noindent
\textbf{HMDB-51} \cite{Kuehne11}: This dataset consists of 6849 videos divided into 51 action classes, with each class containing a minimum of 101 videos. The actions categories can be grouped in five types:
\textit{general facial actions like smile, laugh, chew, and talk; facial actions with object manipulation like smoke, eat, and drink; general body movements like cartwheel, clap hands, climb, climb stairs; body movements with object interaction like brush hair, catch, draw sword, dribble;} and \textit{body movements for human interaction like fencing, hug, kick someone, kiss}.
This is also a challenging dataset as the use of video clips extracted from real-world videos possess the presence of significant camera/background motion alongside varying illumination.

\subsection{Contribution of Static and Motion Domains}

The derivation done in Cholesky based method for fusing the static and motion vectors,
provides us an insightful intuition: we can control the contribution
of motion and static domains to the fusion vector by varying the $\rho$ value.
The derivation of $\rho$ values for different contribution ratios is illustrated in
table \ref{tbl:rho change}.

\begin{table}
\caption{Derivation of $\rho$ values for different contribution levels of static and motion domains to the fused vector}\label{ta:table1}
\begin{center}
  \begin{tabular}{ | c | c | c | }
    \hline
    Contribution to $Z$ & $\rho$ value & Fusion vector \\ \hline \hline

    {\makecell{ 80\% Motion, 20\% Static \\ $\rho_{1}=4\rho_{2}$ }} & \makecell{$\frac{1}{4}\rho_{1}=\sqrt{1-\rho_{1}^2}$ \\ $\rho_{1} = \frac{4}{\sqrt{17}}$} & $Z=\frac{4}{\sqrt{17}}M + \frac{1}{\sqrt{17}}S$ \\ \hline

    {\makecell{ 60\% Motion, 40\% Static \\ $2\rho_{1}=3\rho_{2}$ }} & \makecell{$\frac{2}{3}\rho_{1}=\sqrt{1-\rho_{1}^2}$ \\ $\rho_{1} = \frac{3}{\sqrt{13}}$} & $Z=\frac{3}{\sqrt{13}}M + \frac{2}{\sqrt{13}}S$ \\ \hline

      {\makecell{ 50\% Motion, 50\% Static \\ $\rho_{1}=\rho_{2}$ }} & \makecell{$\rho_{1}=\sqrt{1-\rho_{1}^2}$ \\ $\rho_{1} = \frac{1}{\sqrt{2}}$} & $Z=\frac{1}{\sqrt{2}}M + \frac{1}{\sqrt{2}}S$ \\ \hline

    {\makecell{ 40\% Motion, 60\% Static \\ $3\rho_{1}=2\rho_{2}$ }} & \makecell{$\frac{3}{2}\rho_{1}=\sqrt{1-\rho_{1}^2}$ \\ $\rho_{1} = \frac{2}{\sqrt{13}}$} & $Z=\frac{2}{\sqrt{13}}M + \frac{3}{\sqrt{13}}S$ \\ \hline

    \makecell{80\% Motion, 20\% Static \\ $4\rho_{1}=\rho_{2}$} & \makecell{$4\rho_{1}=\sqrt{1-\rho_{1}^2}$ \\ $\rho_{1} = \frac{1}{\sqrt{17}}$} & {$Z=\frac{1}{\sqrt{17}}M + \frac{4}{\sqrt{17}}S$} \\ \hline

      \label{tbl:rho change}
  \end{tabular}
\end{center}

\end{table}

Results for these different contribution values for UCF-11 and Hollywood2 datasets, are shown in table \ref{tbl:rho ucf} and table \ref{tbl:rho hollywood2}.
We use accuracy and mean average precision as performance metrics, for UCF-11 and
Hollywood2, respectively. For both datasets, we obtain the optimum contribution ratio as 80:20 between static and motion vectors. In the case of HMDB51 dataset, the optimum contribution ration is obtained as 60:40 between static and motion vectors.

\begin{table}[]
\centering
\caption{Overall accuracy of UCF-11, Hollywood2, and HMDB51 for varying ratios between static and motion components. The vectors are fused using Cholesky method. Ratios are indicated in the
format static:motion.} \label{tbl:compare_all}
\begin{tabular}{|l||c|c|c|}
\hline
Dataset           & UCF-11      & Hollywood2       & HMDB51    \\ \hline  \hline
100:0             & 91.8\%      & 56.9\%           & 48.2\%    \\
80:20     & \textbf{96.3\%}     & \textbf{80.9\%}  & 62.25\%   \\
60:40             & 95.3\%      & 73.6\%    &\textbf{67.24\%}  \\
50:50             & 95.3\%      & 64.9\%           & 58.64\%   \\
40:60             & 93.6\%      & 60.3\%           & 42.48\%   \\
20:80             & 91.8\%      & 51.9\%           & 40.43\%   \\ \hline
\end{tabular}
\end{table}

\begin{table*}[]
\centering
\caption{Per-class accuracy for different contribution of static and motion vectors for UCF-11. The vectors are fused using Cholesky method. Ratios are indicated in the
format static:motion. Highest accuracy for UCF-11 is achieved using a
80:20 ratio between static and motion vectors.}\label{tbl:rho ucf}
\begin{tabular}{|l||l|l|l|l|l|l|l|}
\hline
Class           & 100:0  & 80:20     & 60:40   & 50:50    & 40:60     & 20:80   & 0:100 \\ \hline  \hline
B\_shooting      & 92.4\% & \textbf{96.3\%}   & 92.7\%  &  96.3\%  &  91.3\%   & 91.9\%  & 91.3\%  \\
Biking          & 94.3\% &  \textbf{97.8\%}   & 95.6\%  &  95.4\%  &  95.4\%   & 92.6\%  & 89.5\%   \\
Diving          & 90.3\% &  \textbf{95.8\%}   & 94.3\%  &  94.3\%  &  93.1\%   &  89.6\% & 86.2\%  \\
G\_swinging      & 93.2\% & \textbf{96.7\%}   &  96.0\% &  95.8\%  &  93.3\%   &  92.8\% & 90.5\% \\
H\_riding        & 94.0\% &  \textbf{98.0\%}   &  96.6\% &  95.6\%  &  93.1\%   &  90.2\% & 87.2\% \\
S\_juggling      & 92.4\%&  \textbf{96.5\%}    & 96.0\%  &  96.0\%  &  93.7\%   &  90.2\% & 85.4\%  \\
Swinging        & 89.3\%&  \textbf{94.3\%}    & 94.3\%  &  93.6\%  &  94.1\%   &  91.7\% & 88.2\% \\
T\_swinging      & 92.3\%& \textbf{96.9\% }    &  95.7\% &  94.5\%  &  94.1\%   &  93.3\% &  90.6\% \\
T\_jumping       & 93.7\%&  \textbf{97.6\%}    &  96.7\% &  94.5\%  &  94.1\%   &  93.1\% & 90.6\% \\
V\_spiking       & 88.2\%&  93.4\%    &  94.2\% &  \textbf{97.2\%}  &  94.1\%   &  93.0\% & 89.3\% \\
W\_dog           & 90.2\%&  \textbf{96.7\% }   &   96.2\%&  95.4\%  &  93.3\%   &  91.9\% & 87.2\% \\  \hline
Accuracy        &   91.8\%&  \textbf{96.3\%}    &   95.3\%&  95.3\%  &  93.6\%   & 91.8\%  & 88.72\%  \\ \hline
\end{tabular}
\end{table*}

\begin{table*}[]
\centering
\caption{mAP for each class for different contribution of static and motion vectors to the fused vector for Hollywood2. ratios are indicated in the
format static:motion. Highest mAP for Hollywood2 is achieved using a
80:20 ratio between static and motion vectors.}\label{tbl:rho hollywood2}
\begin{tabular}{|l||l|l|l|l|l|l|l|}
\hline
Class           & 100:0 & 80:20 & 60:40 & 50:50 & 40:60 & 20:80 & 0:100 \\ \hline \hline
AnswerPhone     & 52.3\%& \textbf{76.6\%}& 49.6\%& 42.4\%& 38.2\%& 36.6\%& 28.2\%        \\
DriveCar        & 54.6\%& \textbf{98.1\%}& 49.2\%& 42.5\%& 39.1\%& 37.7\%& 26.4\%            \\
Eat             & 50.0\%& \textbf{62.1\%}& 55.6\%& 53.2\%& 53.2\%& 50.1\%& 40.0\%           \\
FightPerson     & 72.2\%& \textbf{94.3\%}& 80.2\%& 72.8\%& 66.6\%& 57.4\%& 42.2\%  \\
GetOutCar       & 56.9\%& \textbf{77.4\%}& 56.2\%& 50.2\%& 47.3\%& 40.2\%& 35.5\%\\
HandShake       & 42.2\%& 78.9\%& \textbf{80.2\%}& 72.7\%& 64.4\%& 50.3\%& 42.7\% \\
HugPerson       & 49.9\%& \textbf{77.1\%}& 64.3\%& 62.6\%& 57.2\%& 50.9\%& 40.6\% \\
Kiss            & 49.9\%& 85.3\%& \textbf{86.4\%}& 70.2\%& 68.7\%& 60.8\%& 45.5\% \\
Run             & 60.2\%& 78.2\%& \textbf{94.8\%}& 88.2\%& 82.3\%& 72.2\%& 64.5\%  \\
SitDown         & 80.2\%& 86.2\%& \textbf{91.6\%}& 80.4\%& 76.9\%& 67.3\%& 56.9\%  \\
SitUp           & 58.7\%& 75.0\%& \textbf{78.2\%}& 70.2\%& 67.4\%& 51.8\%& 47.2\% \\
StandUp         & 55.5\%& 81.2\%& \textbf{97.4\%}& 73.6\%& 62.1\%& 48.3\%& 32.3\% \\ \hline
mAP             &  56.9\% & 80.9\%   &  73.6\%    &    64.9\%   &    60.3\%   &   51.9\%    &  41.8\%\\ \hline

\end{tabular}
\end{table*}

\begin{figure}
\centering
\begin{tikzpicture}
\begin{axis}[
    title={},
    xlabel={Contribution level of motion and static domains},
    ylabel={Accuracy},
    ymin=80, ymax=100,
    symbolic x coords={100:0,80:20,60:40,50:50,40:60,20:80,0:100},
    ytick={80,85,90,95,100,105,110},
    legend pos=north west,
    ymajorgrids=true,
    label style={font=\small},
    grid style=dashed,
    legend style={at={(0.5,-0.2)},anchor=north}
]

    \addplot[
    color=blue,
    mark=square,
    ]
    coordinates {
    (100:0,92.4)(80:20,96.3)(60:40,92.7)(50:50,96.3)(40:60,91.3)(20:80,91.9)(0:100,91.3)
    };
    \addlegendentry{B shooting}

    \addplot[
    color=red,
    mark=square,
    ]
    coordinates {
    (100:0,94.3)(80:20,97.8)(60:40,95.6)(50:50,95.4)(40:60,95.4)(20:80,92.6)(0:100,89.5)
    };
    \addlegendentry{Biking}

\addplot[
    color=cyan,
    mark=square,
    ]
    coordinates {
    (100:0,90.3)(80:20,95.8)(60:40,94.3)(50:50,94.3)(40:60,93.1)(20:80,89.6)(0:100,86.2)
    };
    \addlegendentry{Diving}

\addplot[
    color=green,
    mark=square,
    ]
    coordinates {
    (100:0,  93.2)(80:20, 96.7 )(60:40, 96.0 )(50:50,95.8  )(40:60,93.3  )(20:80, 92.8 )(0:100, 90.5 )
    };
    \addlegendentry{G swinging}

\addplot[
    color=magenta,
    mark=square,
    ]
    coordinates {
    (100:0, 94.0 )(80:20,  98.0)(60:40, 96.6 )(50:50,95.6  )(40:60, 93.1 )(20:80,90.2  )(0:100, 87.2 )
    };
    \addlegendentry{H riding}

\addplot[
    color=black,
    mark=square,
    ]
    coordinates {
    (100:0,92.4  )(80:20, 96.5 )(60:40, 96.0 )(50:50,96.0  )(40:60, 93.7 )(20:80, 90.2 )(0:100, 85.4 )
    };
    \addlegendentry{S juggling}

\addplot[
    color=lime,
    mark=square,
    ]
    coordinates {
    (100:0, 89.3 )(80:20, 94.3 )(60:40,94.3  )(50:50, 93.6 )(40:60, 94.1 )(20:80, 91.7 )(0:100, 88.2 )
    };
    \addlegendentry{ Swinging   }

\addplot[
    color=yellow,
    mark=square,
    ]
    coordinates {
    (100:0, 92.3 )(80:20,96.9  )(60:40, 95.7 )(50:50, 94.5 )(40:60, 94.1 )(20:80, 93.3 )(0:100, 90.6 )
    };
    \addlegendentry{  T swinging  }

\addplot[
    color=olive,
    mark=square,
    ]
    coordinates {
    (100:0, 93.7 )(80:20, 97.6 )(60:40, 96.7 )(50:50, 94.5 )(40:60,94.1  )(20:80, 93.1 )(0:100, 90.6 )
    };
    \addlegendentry{  T jumping  }

\addplot[
    color=gray,
    mark=square,
    ]
    coordinates {
    (100:0, 88.2 )(80:20, 93.4 )(60:40, 94.2 )(50:50, 97.2 )(40:60, 94.1 )(20:80, 93.0 )(0:100,  89.3 )
    };
    \addlegendentry{   V spiking }

\addplot[
    color=orange,
    mark=square,
    ]
    coordinates {
    (100:0, 90.2 )(80:20, 96.7 )(60:40,96.2  )(50:50,95.4  )(40:60,  93.3)(20:80, 91.9 )(0:100, 87.2 )
    };
    \addlegendentry{  W dog  }

\end{axis}
\end{tikzpicture}
\caption{Accuracy distribution for different contribution levels of motion and static domains.
This figure illustrates that motion:static ratio affects the accuracy and the optimum contribution depends on the UCF-11 dataset for each class.}
\label{contribution chart ucf}
\end{figure}
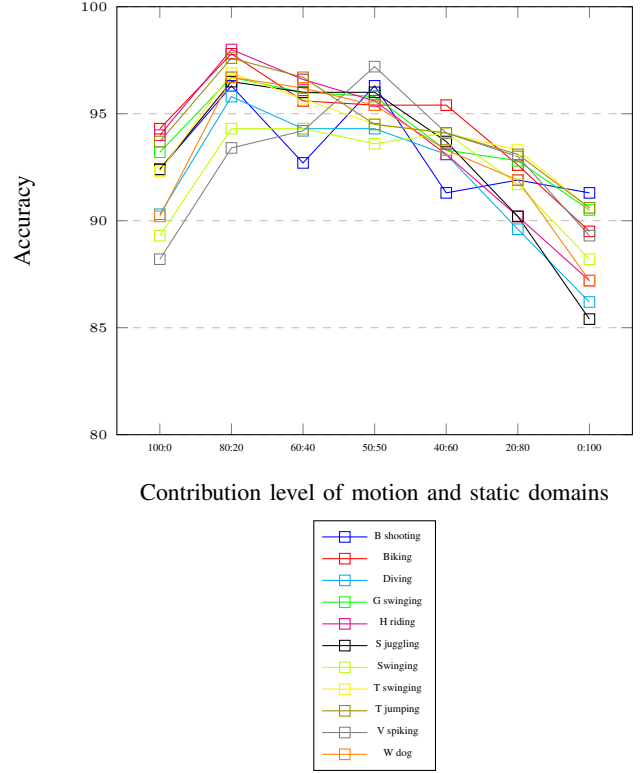

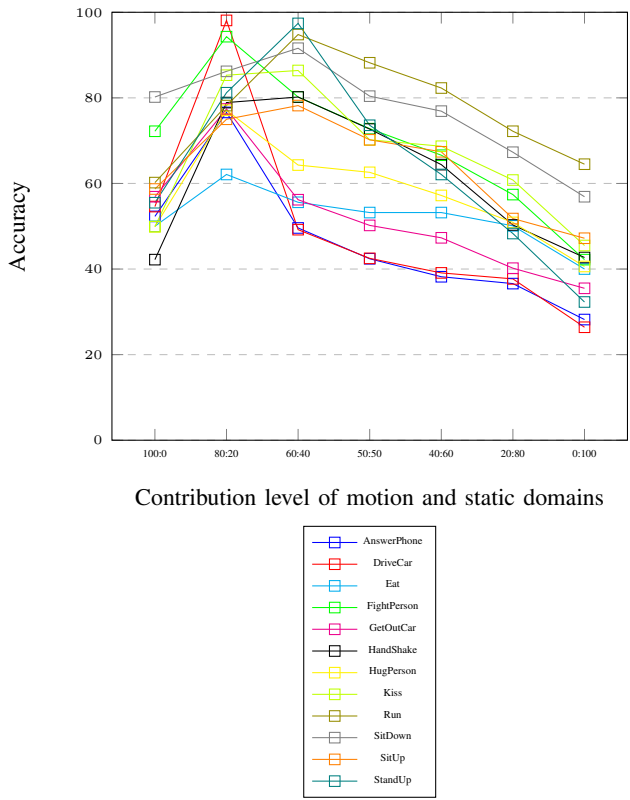
\begin{figure}
\centering
\begin{tikzpicture}
\begin{axis}[
    title={},
    xlabel={Contribution level of motion and static domains},
    ylabel={Accuracy},
    ymin=0, ymax=100,
    symbolic x coords={100:0,80:20,60:40,50:50,40:60,20:80,0:100},
    ytick={0,20,40,60,80,100,120},
    legend pos=north west,
    ymajorgrids=true,
    label style={font=\small},
    grid style=dashed,
    legend style={at={(0.5,-0.2)},anchor=north}
]

\addplot[
    color=blue,
    mark=square,
    ]
    coordinates {
    (100:0, 52.3 )(80:20, 76.6 )(60:40, 49.6  )(50:50, 42.4   )(40:60,   38.2 )(20:80,   36.6   )(0:100,  28.2   )
    };
    \addlegendentry{    AnswerPhone    }

\addplot[
    color=red,
    mark=square,
    ]
    coordinates {
    (100:0, 54.6 )(80:20,98.1  )(60:40, 49.2  )(50:50, 42.5   )(40:60, 39.1   )(20:80,   37.7   )(0:100,  26.4   )
    };
    \addlegendentry{    DriveCar    }

\addplot[
    color=cyan,
    mark=square,
    ]
    coordinates {
    (100:0,50.0  )(80:20, 62.1 )(60:40, 55.6  )(50:50,  53.2  )(40:60,  53.2  )(20:80,   50.1   )(0:100,  40.0   )
    };
    \addlegendentry{    Eat    }

\addplot[
    color=green,
    mark=square,
    ]
    coordinates {
    (100:0, 72.2 )(80:20, 94.3 )(60:40, 80.2  )(50:50,  72.8  )(40:60,  66.6  )(20:80,   57.4   )(0:100,   42.2  )
    };
    \addlegendentry{    FightPerson    }

\addplot[
    color=magenta,
    mark=square,
    ]
    coordinates {
    (100:0, 56.9 )(80:20, 77.4 )(60:40,  56.2 )(50:50, 50.2   )(40:60,  47.3  )(20:80,   40.2   )(0:100,  35.5   )
    };
    \addlegendentry{   GetOutCar     }

\addplot[
    color=black,
    mark=square,
    ]
    coordinates {
    (100:0, 42.2 )(80:20, 78.9 )(60:40,  80.2 )(50:50,  72.7  )(40:60, 64.4   )(20:80,  50.3    )(0:100,   42.7  )
    };
    \addlegendentry{     HandShake   }

\addplot[
    color=yellow,
    mark=square,
    ]
    coordinates {
    (100:0, 49.9 )(80:20, 77.1 )(60:40, 64.3  )(50:50,  62.6  )(40:60,  57.2  )(20:80,  50.9   )(0:100,   40.6  )
    };
    \addlegendentry{    HugPerson    }

\addplot[
    color=lime,
    mark=square,
    ]
    coordinates {
    (100:0,49.9  )(80:20, 85.3 )(60:40, 86.4  )(50:50, 70.2   )(40:60,  68.7  )(20:80,  60.8    )(0:100,   45.5  )
    };
    \addlegendentry{     Kiss   }

\addplot[
    color=olive,
    mark=square,
    ]
    coordinates {
    (100:0, 60.2 )(80:20, 78.2 )(60:40,94.8   )(50:50,  88.2  )(40:60, 82.3   )(20:80,   72.2   )(0:100, 64.5    )
    };
    \addlegendentry{     Run   }

\addplot[
    color=gray,
    mark=square,
    ]
    coordinates {
    (100:0, 80.2 )(80:20, 86.2 )(60:40, 91.6  )(50:50,  80.4  )(40:60, 76.9   )(20:80,  67.3    )(0:100,   56.9  )
    };
    \addlegendentry{    SitDown    }

\addplot[
    color=orange,
    mark=square,
    ]
    coordinates {
    (100:0, 58.7 )(80:20,75.0  )(60:40,  78.2 )(50:50, 70.2   )(40:60,   67.4 )(20:80,  51.8    )(0:100,  47.2   )
    };
    \addlegendentry{    SitUp    }

\addplot[
    color=teal,
    mark=square,
    ]
    coordinates {
    (100:0, 55.5 )(80:20,81.2  )(60:40,  97.4 )(50:50,  73.6  )(40:60,    62.1)(20:80, 48.3     )(0:100, 32.3    )
    };
    \addlegendentry{    StandUp    }

\end{axis}
\end{tikzpicture}
\caption{Accuracy distribution for different contribution levels of motion and static domains.
This figure illustrates that motion:static ratio affects the accuracy and the optimum contribution depends on the Hollywood2 dataset.}
\label{contribution chart hollywood}
\end{figure}

An overview distribution of the overall performance over different contribution levels, from static and motion domains, for both datasets is shown in Fig. \ref{contribution chart ucf}
and Fig. \ref{contribution chart hollywood}.
We can see that the performance change for different contribution percentages of motion and static domain. Also, the optimum contribution may change
depending on the nature of the action and richness of motion or static information in the video. For example, if the motion patterns are indistinguishable across actions, static information plays
a critical role, for determining the action, and vise versa. The amount of interaction with objects also may play a key role in determining this ratio. In Fig. \ref{contribution chart hollywood} it is evident that for action classes which does not
highly interact with external objects---such as \textit{kiss,run, sitdown, situp, standup handshake}---, the optimum motion:static ratio is 40:60. For other action classes, which
interact with objects, optimum motion:static ratio is 20:80. This highlights
our hypothesis, that being able to control this contribution explicitly, is vital for an action recognition system.

\pgfplotstableread[row sep=\\,col sep=&]{
    class     &   & RF  \\
    Bshooting   & 96.3  & 90.2   \\
    Biking       & 97.8 & 93.2    \\
    Diving       & 95.8 & 93.2 \\
    Gswinging   & 96.7 & 90.3  \\
    Hriding     & 98.0  & 92.3 \\
    Sjuggling   & 96.5  & 93.8  \\
    Swinging     & 94.3  & 89.2  \\
    Tswinging   & 96.9  & 91.3  \\
    Tjumping    & 97.6  & 94.5  \\
    Vspiking    & 93.4  & 89.6  \\
    Wdog        & 96.7  & 91.9  \\
    Accuracy     & 96.3  & 84.32  \\
    }\mydata

\subsection{Mathematical Validation of Optimum Contribution}

As it is evident from the results of table \ref{tbl:rho ucf} and table \ref{tbl:rho hollywood2}, we experimentally obtain the
optimum contribution ratio for the datasets UCF-11 and Holllywood2 as 80:20. In section III, in the derivation of the variance ratio based fusion model,
we mathematically obtained values for the optimum contribution as 70:30 for the same datasets. It should be noted that these values closely
represent the experimental values, and hence, the results are further verified.

\subsection{Comparison of Fusion Models}
The per-class accuracies obtained for each fusion model is illustrated in table \ref{tbl:per-action fusionucf} and
table \ref{tbl:per-action fusionhollywood}. Although all three methods give impressive results, Cholesky based fusion
model is superior, and has an overall accuracy of 96.3\% for UCF-11. For Hollywood2 dataset, it achieves a mean average precision
of 80.9\%.

\begin{table}[]
\centering
\caption{Comparison of fusion models on UCF-11 dataset.}\label{tbl:per-action fusionucf}
\begin{tabular}{|l||l|l|l|l|l|}
\hline
Class            & Cholesky & Variance ratio & PCA   \\ \hline  \hline
B\_shooting       & \textbf{96.3\%}    &  90.3\%   &  90.6\%  \\
Biking           & \textbf{97.8\%}    &  90.8\%   &  91.0\%    \\
Diving           & \textbf{95.8\%}    &  92.3\%   &  89.3\%   \\
G\_swinging       & \textbf{96.7\%}    &  90.3\%   &  92.3\%   \\
H\_riding         & \textbf{98.0\%}    &  87.4\%   &  88.6\%    \\
S\_juggling       & \textbf{96.5\%}    &  89.7\%   &  92.8\%    \\
Swinging         & \textbf{94.3\%}    &  90.0\%   &  88.0\%    \\
T\_swinging       & \textbf{96.9\%}    &  89.4\%   &  93.0\%   \\
T\_jumping        & \textbf{97.6\%}    &  92.5\%   &  91.0\%    \\
V\_spiking        &\textbf{93.4\%}    &  91.6\%   &  91.7\%   \\
W\_dog            & \textbf{96.7\%}    &  91.6\%   &  93.4\%   \\ \hline \hline
Accuracy &  96.3\%   &  90.5\%   &   91.1\%   \\ \hline
\end{tabular}
\end{table}

\begin{table}[]
\centering
\caption{Comparison of fusion models on Hollywood2 dataset}\label{tbl:per-action fusionhollywood}
\begin{tabular}{|l||l|l|l|l|l|}
\hline
Class            & Cholesky & Variance ratio & PCA   \\ \hline  \hline
AnswerPhone      & \textbf{76.6\%}& 62.4\%    & 67.6\%           \\
DriveCar         & \textbf{98.1\%}& 72.8\%    & 70.0\%       \\
Eat              & \textbf{62.1\%}& 49.4\%    & 56.5\%          \\
FightPerson      & \textbf{94.3\%}& 78.2\%    & 72.6\%         \\
GetOutCar        & \textbf{77.4\%}& 46.9\%    & 56.7\%          \\
HandShake        & \textbf{78.9\%}         & 56.9\%    & 55.6\%            \\
HugPerson        & \textbf{77.1\%}& 52.4\%    & 60.6\%            \\
Kiss             &\textbf{85.3\%}         & 64.0\%    & 66.6\%           \\
Run              & \textbf{78.2\%}         & 58.3\%    & 54.3\%            \\
SitDown          & \textbf{86.2\%}         & 72.0\%    & 68.6\%            \\
SitUp            & \textbf{75.0\%}         & 50.0\%    & 54.7\%          \\
StandUp          & \textbf{81.2\%}         & 54.4\%    & 50.0\%          \\ \hline
mAP              &      80.9\%          &   59.8\%        &       61.1\%             \\ \hline
\end{tabular}
\end{table}

\subsection{Comparison with the state-of-the-art}

Table \ref{tbl:comparison} compares our results to state of the art. We use a motion:static ratio of 20:80 for both "change/HMDB" datasets to combine the static and motion vectors,
since these values gave the best results. On UCF-11, we significantly outperform
the state of the art Ramasinghe \emph{et al.}~\cite{7486474} by 3.2\%. A mean average precision of 80.9\% is achieved by our system for Hollywood2, which outperforms
the state-of-the-art by 16.6\%. "add here"

\begin{table*}[]
\centering
\caption{Comparison of our method with state-of-the-art methods in the literature. Static:motion ratios are 80:20 for UCF-11 and Hollywood2, and 60:40 for HMDB51. The results mentioned of our system are those obtained using the Cholesky method.}\label{tbl:comparison}
\begin{tabular}{|l|l||l|l|l|l|}
\hline
\multicolumn{2}{|c||}{UCF-11}    & \multicolumn{2}{c|}{Hollywood2} & \multicolumn{2}{c|}{HMDB51}  \\ \hline
Liu \textit{et al.}\cite{liu2009recognizing}  & 71.2\%   & Vig \textit{et al.}\cite{vig2012space} & 59.4\%   &  Wang \textit{et al.}\cite{wang2013action} & 57.2\%  \\
Ikizler-Cinbis \textit{et al.}\cite{ikizler2010object} & 75.21\% & Jiang \textit{et al.}\cite{jiang2012trajectory}  & 59.5\%  &  Wang \textit{et al.}\cite{wang2015action} & 65.9\%  \\
Wang \textit{et al.}\cite{wang2011action}    & 84.2\%  & Mathe \textit{et al.}\cite{mathe2012dynamic}         & 61.0\%    & Zhu  \textit{et al.}\cite{ZhuN16} & \textbf{68.2\%} \\
Sameera \textit{et al.}\cite{7486474}         & 93.1\%        & Jain \textit{et al.}\cite{jain2013better}           & 62.5\%  & &     \\
                      &         & Wang \textit{et al.}\cite{wang2011action}             & 58.3\%     & & \\
                      &         & Wang \textit{et al.}\cite{wang2013action}          & 64.3\%      & &\\ \hline \hline
Our method  & \textbf{96.3\%}       & Our method        &        \textbf{80.9\%} & Our method & 67.24\%       \\ \hline

\end{tabular}

\end{table*}

\begin{table*}[]
\centering
\caption{Per-class accuracy comparison with state-of-the-art on UCF-11.}\label{tbl:per-action ucf}
\begin{tabular}{|l||l|l|l|l|l|}
\hline
Class            & Ours(Cholesky) & KLT\cite{lucas1981iterative} & Wang et al.\cite{wang2011action} & Ikizler-Cinbis\cite{ikizler2010object} & Ramasinghe et al.\cite{7486474} \\ \hline  \hline
B\_shooting       & \textbf{96.3\%}    &  34.0\%   &  43.0\%   & 48.5\%    &   95.6\%  \\
Biking           & \textbf{97.8\%}    &  87.6\%   &  91.7\%   & 75.17\%    &  93.1\%   \\
Diving           & 95.8\%    &  \textbf{99.0\%}   &  \textbf{99.0\%}   & 95.0\%    &   92.8\%  \\
G\_swinging       & 96.7\%    &  95.0\%   &  \textbf{97.0\%}   & 95.0\%    &   95.0\%  \\
H\_riding         & \textbf{98.0\%}    &  76.0\%   &  85.0\%   & 73.0\%    &   94.3\%  \\
S\_juggling       & \textbf{96.5\%}    &  65.0\%   &  76.0\%   & 53.0\%    &   87.8\%  \\
S\_winging         & \textbf{94.3\%}    &  86.0\%   &  88.0\%   & 66.0\%    &   92.4\%  \\
T\_swinging       & \textbf{96.9\%}    &  71.0\%   &  71.0\%   & 77.0\%    &   94.9\%  \\
T\_jumping        & \textbf{97.6\%}    &  93.0\%   &  94.0\%   & 93.0\%    &   94.0\%  \\
V\_spiking        & 93.4\%    &  \textbf{96.0\%}   &  95.0\%   & 85.0\%    &   93.2\%  \\
W\_dog            & \textbf{96.7\%}    &  76.4\%   &  87.0\%   & 66.7\%    &   91.4\%  \\ \hline \hline
Accuracy &  \textbf{96.3\%}   &  79.0\%   &  84.2\%   & 75.2\%    &   93.1\%  \\ \hline
\end{tabular}
\end{table*}

\begin{table*}[]
\centering
\caption{Per-class mAP comparison with state-of-the-art on Hollywood2.}\label{tbl:per-action hollywood}
\begin{tabular}{|l|l|l|l|l|l|}
\hline
Class            & Ours           & KLT\cite{lucas1981iterative} & Wang et al.\cite{wang2011action} & Ullah\cite{ullah2010improving}   \\ \hline \hline
AnswerPhone      & \textbf{76.6\%}& 18.3\%    & 32.6\%    & 25.9\%        \\
DriveCar         & \textbf{98.1\%}& 88.8\%    & 88.0\%    & 85.9\%     \\
Eat              & \textbf{62.1\%}& 73.4\%    & 65.2\%    & 56.4\%         \\
FightPerson      & \textbf{94.3\%}& 74.2\%    & 81.4\%    & 74.9\%         \\
GetOutCar        & \textbf{77.4\%}& 47.9\%    & 52.7\%    & 44.0\%       \\
HandShake        & 78.9\%         & 18.4\%    & 29.6\%    & 29.7\%         \\
HugPerson        & \textbf{77.1\%}& 42.6\%    & 54.2\%    & 46.1\%         \\
Kiss             & 85.3\%         & 65.0\%    & 65.8\%    & 55.0\%         \\
Run              & 78.2\%         & 76.3\%    & 82.1\%    & 69.4\%         \\
SitDown          & 86.2\%         & 59.0\%    & 62.5\%    & 58.9\%         \\
SitUp            & 75.0\%         & 27.7\%    & 20.0\%    & 18.4\%         \\
StandUp          & 81.2\%         & 63.4\%    & 65.2\%    & 57.4\%         \\ \hline
mAP              &     \textbf{80.9\%}           &   54.6\%         &     58.3\%       &     51.8\%      \\ \hline
\end{tabular}
\end{table*}

Per-action class results, are also compared in table \ref{tbl:per-action ucf} and table \ref{tbl:per-action hollywood}. In UCF-11, our method excells
in 8 out of 11 classes, when compared with Lucas et al.\cite{lucas1981iterative}, Wang et al.\cite{wang2011action}, Ikizler et al.\cite{ikizler2010object}
and Ramasinghe \emph{et al.}~\cite{7486474}. In Hollywood2,
we calculate the average precision of each class, and compare with Lucas et al.\cite{lucas1981iterative}, Wang et al.\cite{wang2011action}, and \cite{ullah2010improving}.
We achieve best results in 10 out of 12 classes in this case.

\subsection{Effectiveness of Capturing Time Evolution}
\begin{figure}
  \centering
   \begin{tikzpicture}
    \begin{axis}[
            ybar,
            bar width=.1cm,
            width=0.5\textwidth,
            height=.2\textwidth,
            legend pos=north west,
            label style={font=\small},
    ymajorgrids=true,
    grid style=dashed,
    legend style={at={(0.5,-0.5)},anchor=north},
            symbolic x coords={B\_shooting,Biking,Diving,G\_swinging,H\_riding,S\_juggling,Swinging,T\_swinging,T\_jumping,V\_spiking,W\_dog,Accuracy},
            x tick label style={rotate=45, anchor=north east, inner sep=0mm},
            xtick=data,
            ymin=0,ymax=100,
            xlabel={Action Classes},
            ylabel={Accuracy},
        ]
        \addplot table[x=class,y=LSTM]{\mydata};
        \addplot table[x=class,y=RF]{\mydata};
        \legend{LSTM, Random Forest}
    \end{axis}
\end{tikzpicture}
\caption{Accuracy comparison between Random Forest Classifier and LSTM for UCF-11 dataset. Motion:static ratio of 20:80 is used. Accuracy is significantly higher
when the temporal dynamics of sub events are captured.}
\label{randomucf}
\end{figure}
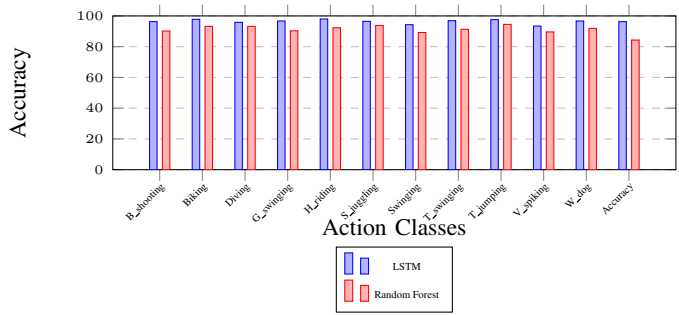

\begin{figure}
  \centering
   \begin{tikzpicture}
    \begin{axis}[
            ybar,
            bar width=.1cm,
            width=0.5\textwidth,
            height=.2\textwidth,
            legend pos=north west,
            label style={font=\small},
    ymajorgrids=true,
    grid style=dashed,
    legend style={at={(0.5,-0.5)},anchor=north},
            symbolic x coords={AnswerPhone,DriveCar,Eat,FightPerson,GetOutCar,HandShake,HugPerson,Kiss,Run,SitDown,SitUp,StandUp,mAP},
            x tick label style={rotate=45, anchor=north east, inner sep=0mm},
            xtick=data,
            ymin=0,ymax=100,
            xlabel={Action Classes},
            ylabel={Accuracy},
        ]
        \addplot table[x=class,y=LSTM]{\mydata};
        \addplot table[x=class,y=RF]{\mydata};
        \legend{LSTM, Random Forest}
    \end{axis}
\end{tikzpicture}
\caption{mAP comparison for Random Forest Classifier and LSTM for Hollywood2 dataset. Motion:static ratio of 20:80 is used. mAP is significantly higher
when the temporal dynamics of sub events are captured.}
\label{randomHollywood}
\end{figure}
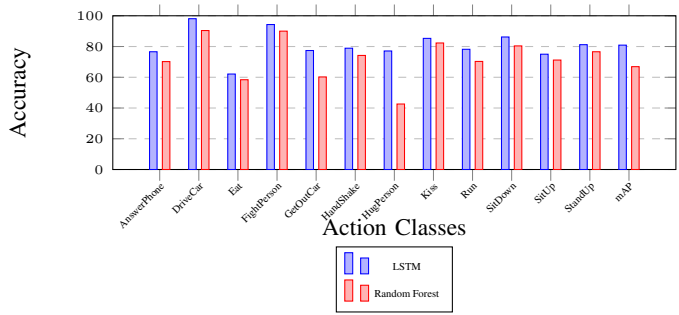

As discussed in earlier sections, complex actions are composed of sub activities preserving
a temporal pattern. In this work, we try to capture those underlying patterns by an LSTM network.
It is interesting to verify whether this strategy has an impact on the accuracy of the
classification. Here we directly feed the fused vectors to a random forest classifier, which
does not capture sequential dynamic patterns, and compare it with the results obtained by
the LSTM network. The results are shown in Fig. \ref{randomucf} and Fig. \ref{randomHollywood}.

As it is evident from the results in in Fig. \ref{randomucf} and Fig. \ref{randomHollywood}, LSTM network significantly outperforms the
random forest classifier for both datasets. In Hollywood2, the LSTM network wins by a 14\% margin. In UCF-11, the LSTM network wins by a 12\% margin.
Therefore, it can be concluded that, exploiting
temporal patterns of sub activities, benefits complex action classification.

\pgfplotstableread[row sep=\\,col sep=&]{
    class     & LSTM & RF  \\
    B\_shooting   & 96.3  & 90.2   \\
    Biking       & 97.8 & 93.2    \\
    Diving       & 95.8 & 93.2 \\
    G\_swinging   & 96.7 & 90.3  \\
    H\_riding     & 98.0  & 92.3 \\
    S\_juggling   & 96.5  & 93.8  \\
    Swinging     & 94.3  & 89.2  \\
    T\_swinging   & 96.9  & 91.3  \\
    T\_jumping    & 97.6  & 94.5  \\
    V\_spiking    & 93.4  & 89.6  \\
    W\_dog        & 96.7  & 91.9  \\
    Accuracy     & 96.3  & 84.32  \\
    }\mydata

\pgfplotstableread[row sep=\\,col sep=&]{
class     & LSTM & RF  \\
AnswerPhone      & 76.6    & 70.2    \\
DriveCar         & 98.1   & 90.4   \\
Eat              & 62.1 & 58.4       \\
FightPerson      & 94.3 &  90.0    \\
GetOutCar        & 77.4  & 60.2  \\
HandShake        & 78.9          & 74.2     \\
HugPerson        & 77.1   & 42.6      \\
Kiss             & 85.3          & 82.3     \\
Run              & 78.2          & 70.3     \\
SitDown          & 86.2          & 80.4     \\
SitUp            & 75.0          & 71.2      \\
StandUp          & 81.2          & 76.6      \\
mAP              & 80.9               & 66.9                  \\
    }\mydata

\section{Conclusion}
This paper presents an end to end system for action classification which operates
on both static and motion features. Our approach relies on deep features,
for creating static vectors, and \textit{motion tubes} for motion features.
Motion tubes are a novel concept we introduce in this paper which can be
used to track individual actors or objects across frames, and model micro level actions.
We present three novel methods: Based on Cholesky transformation, variance ratio, and PCA for efficient combining of features
from different domains, which is a vital requirement in action classification.
Cholesky method provides the power to control the contribution of each domain in exact numbers, and
variance ratio based method mathematically provides an optimum ratio for contribution. We show that these mathematical
and experimental values agree with each other. We run experiments to show that the accuracy depends on the ratio of this contribution, and the optimum contribution of
static and motion domains may vary depending on the richness of motion information. In short, our work demonstrates how the action classification accuracy varies with the combination ratio of static and motion features, while establishing the existence of an optimum combination ratio for a given test dataset.

In addition, we note that this variation of optimum ratio between static and motion feature contribution towards the final classification accuracy being dependent on the dataset demonstrates the possibility of this optimum ratio varying dynamically over time for even a given single video clip. Thus, based on the temporal variation of motion richness of any given video clip, the optimum contribution ratio has a possibility of varying dynamically. We hope to work on this in future.

Through our experiments we also show that our static and motion features are complementary,
and contribute to the final result. We also compare our three fusion algorithms, and
show that the Cholesky based method is superior, although all three of them give impressive results. We also model the temporal progression of sub-events using an LSTM network. Experimental
results indicate that this is indeed beneficial, compared to using models which does not capture temporal dynamics. Comparison of our work with multiple state-of-the-art algorithms, on the  popular datasets, UCF-11, and Hollywood2, show that our system performs better. The work on the HMDB51 dataset shows our system to be on par with the state-of-the-art.

In the future, it would be interesting to improve the motion tubes, so that, it can maintain an identity over each actor object. While it is mostly the case even in the present system, there is no guarantee.
Also, in this work the emphasis is on accelerating the per-actor micro action generation; initially we detect individual objects in first frame, and subsequently track those along motion tubes in the following frames. In the future, exploring more powerful methods to describe micro actions inside motion tubes would be interesting, since it may increase the distinctiveness of the motion features and contribute well to the final accuracy.

\ifCLASSOPTIONpeerreview
\else
\section*{Acknowledgment}
The authors would like to thank the National Research Council of Sri Lanka for funding this research under the Grant 12-018.
\fi

\ifCLASSOPTIONcaptionsoff
  \newpage
\fi


\ifCLASSOPTIONpeerreview
\else
\begin{IEEEbiography}[{\includegraphics[width=1in,height=1.25in,clip,keepaspectratio]{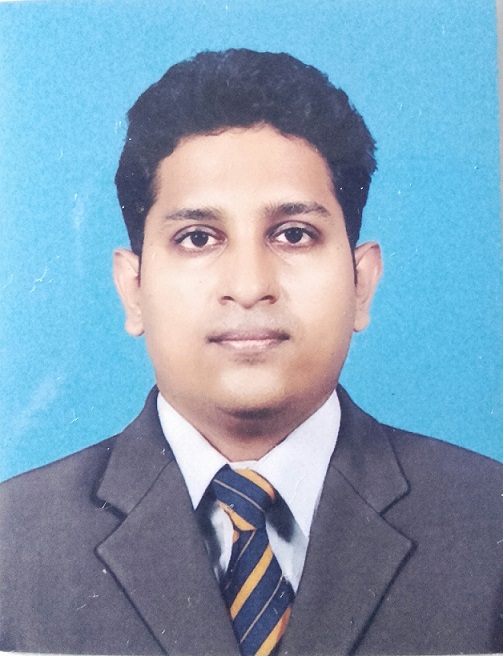}}]{Sameera Ramasinghe}
 obtained the BSc. Engineering degree in Electronics and Telecommunication from University of Moratuwa, Sri Lanka in March 2014. He is a co-founder and a research engineer at ConsientAI, a startup focused on AI technologies. He is currently pursing a M.Phil. at University of Moratuwa. Sri Lanka. His current research interests are machine learning and computer vision.
\end{IEEEbiography}
\begin{IEEEbiography}[{\includegraphics[width=1in,height=1.25in,clip,keepaspectratio]{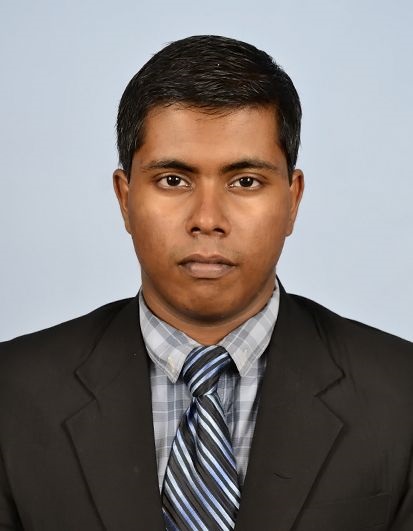}}]{Jathushan Rajasegaran} is currently a 3rd year undergraduate pursuing a B.Sc. Engineering degree in Electronics and Telecommunication from University of Moratuwa, Sri Lanka. His current research interests are machine learning, big data analysis and data privacy.
\end{IEEEbiography}
\begin{IEEEbiography}[{\includegraphics[width=1in,height=1.25in,clip,keepaspectratio]{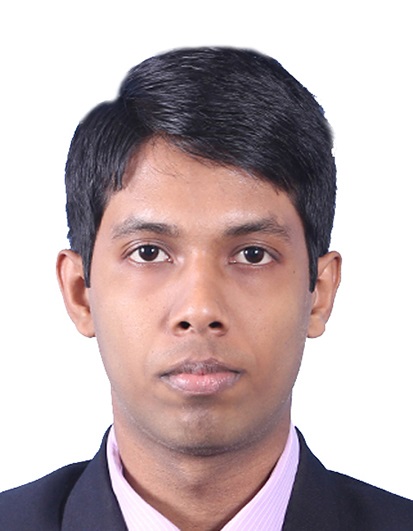}}]{Vinoj Jayasundara} is currently a 3rd year undergraduate pursuing the B.Sc. Engineering (Honours) degree in Electronics and Telecommunication from University of Moratuwa, Sri Lanka. His current research interests are machine learning, big data analytics, machine vision and activity recognition.
\end{IEEEbiography}
\begin{IEEEbiography}[{\includegraphics[width=1in,height=1.25in,clip,keepaspectratio]{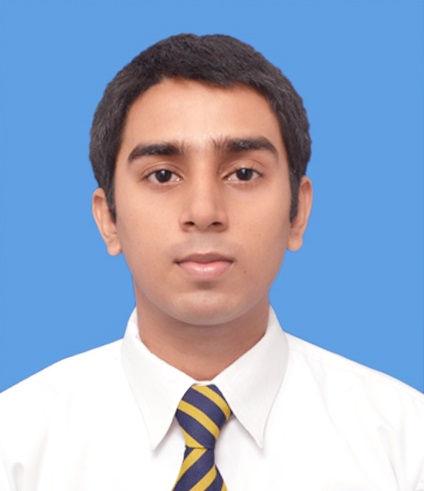}}]{Kanchana Ranasinghe}
 is currently a 2nd year undergraduate pursuing a B.Sc. Engineering degree in Electronics and Telecommunication from University of Moratuwa, Sri Lanka. He will be graduating in 2019. His current research interests are machine learning, computer vision, and pattern recognition.
\end{IEEEbiography}
\begin{IEEEbiography}[{\includegraphics[width=1in,height=1.25in,clip,keepaspectratio]{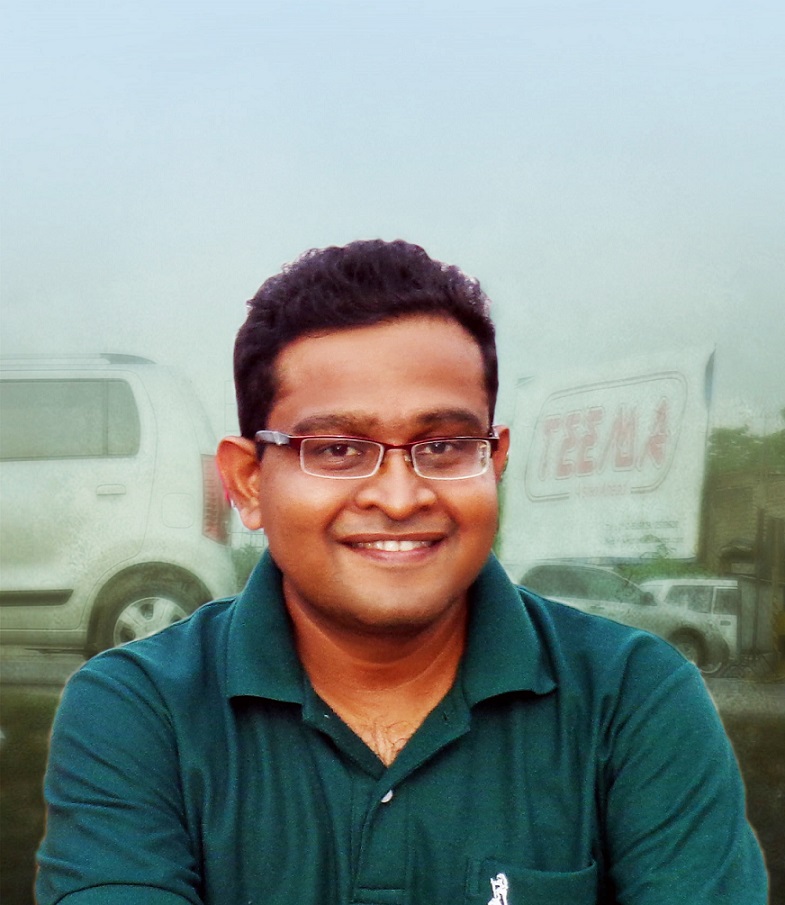}}]{Ranga Rodrigo} received the B.Sc. Eng. Degree (first-class honors) from the University of Moratuwa, Moratuwa, Sri Lanka, in 2001 and the M. E. Sc. and Ph. D. degrees from the Western University, London, ON, Canada in 2004 and 2008, respectively. He has been with the Department of Electronic and Telecommunication Engineering, the University of Moratuwa, since January 2008, where he is a Senior Lecturer. His research interests are in the general area of computer vision. He works in feature tracking, reconstruction, and activity recognition.
\end{IEEEbiography}
\begin{IEEEbiography}[{\includegraphics[width=1in,height=1.25in,clip,keepaspectratio]{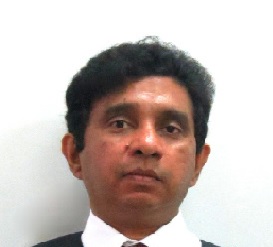}}]{Ajith A. Pasqual} received his B.Sc. Engineering degree with First Class Honours from University of Moratuwa, Sri Lanka in Electronic and Telecommunication Engineering in 1993, M.Eng. and Ph.D. degrees from The University of Tokyo in Computer Vision in 1998 and 2001 respectively. He is currently a Senior Lecturer and was a former Head of Department of Electronic and Telecommunication Engineering, University of Moratuwa.
His primary research interests are in Application processors, Machine Vision, Processor and SoC Architectures and he leads the Reconfigurable Digital Systems Research Group at the University of Moratuwa which work in the area of hardware acceleration, novel architectures for application specific processors and SoCs to improve performance and power efficiency.  He is the founder of the first Semiconductor Startup Company in Sri Lanka – Paraqum Technologies which is developing high performance hardware decoder and encoder for the newest Video Compression Standard – H.265/HEVC and network analytics equipment.

\end{IEEEbiography}
\fi

\end{document}